\documentclass[runningheads]{llncs}
\usepackage[T1]{fontenc}
\usepackage{graphicx}
\usepackage{subfigure}
\usepackage{enumitem}
\usepackage{amssymb}
\usepackage{array}
\usepackage{multicol}
\usepackage{multirow}
\usepackage{pdfpages}
\usepackage{wrapfig}
\usepackage{amsmath}     
\usepackage[colorlinks=true,   
            linkcolor=blue,    
            citecolor=blue,    
            urlcolor=blue      
]{hyperref}
\usepackage{url}            
\usepackage{booktabs}       
\usepackage{amsfonts}       
\usepackage{nicefrac}       
\usepackage{microtype}      
\usepackage{xcolor}         
\usepackage[normalem]{ulem}
\useunder{\uline}{\ul}{}
\usepackage{marvosym}

\begin{document}
\title{STAR: Adaptive Spatial-Temporal Normalization for Unified Microservice Incident Management}
\titlerunning{STAR for Unified Microservice Incident Management}
\author{Xinhua Miao\inst{1} \and
Linyu Zhu\inst{1} \and Bowei Yang\inst{2} \and Zhengong Cai\inst{1}\textsuperscript{\Letter}}
\authorrunning{X. Miao et al.}
%
\institute{School of Software Technology, Zhejiang University, Ningbo, China\\
\email{\{xinhuamiao,zhulinyu2003,cstcaizg\}@zju.edu.cn}
\and
School of Aeronautics and Astronautics, Zhejiang University, Hangzhou, China\\
\email{boweiy@zju.edu.cn}}
\maketitle              
\begin{abstract}
Automated incident management in large-scale microservice systems relies on learning robust representations from multimodal observability data, including metrics, logs, and traces. Although recent self-supervised frameworks enable unified modeling for anomaly detection (AD), failure triage (FT), and root cause localization (RCL), they often struggle with non-stationary temporal dynamics and heterogeneous service dependency structures. In this paper, we propose \textbf{STAR}, a \textbf{S}patial-\textbf{T}emporal \textbf{A}daptive \textbf{R}epresentation learning framework that explicitly addresses these challenges through adaptive normalizations. STAR introduces two tightly coupled mechanisms: Temporal Adaptive Normalization (TAN), which dynamically normalizes multivariate time series using multi-scale temporal context, and Spatial Adaptive Normalization (SAN), which performs structure-aware normalization over service dependency graphs. Unlike prior methods that treat normalization as static or task-agnostic, STAR formulates it as a learnable, context-conditioned transformation aligned with the intrinsic properties of microservice systems. The resulting adaptive representations are integrated into a unified self-supervised framework, enabling end-to-end unsupervised support for AD, FT, and RCL tasks. Extensive experiments on two real-world microservice benchmarks demonstrate that STAR consistently outperforms all state-of-the-art baselines, yielding significant and stable improvements across all three tasks. Our results highlight adaptive normalization as a principled and effective mechanism for robust multimodal representation learning in complex software systems.

\end{abstract}

\section{Introduction}
Microservice architectures are widely adopted in cloud-native applications but introduce operational complexity due to non-stationary workloads, evolving topologies, and heterogeneous interactions. Automated incident management—including anomaly detection (AD), failure triage (FT), and root cause localization (RCL)—is crucial for system reliability.

Prior work either addresses tasks separately (Hades~\cite{Hades}, MicroCBR~\cite{MicroCBR}, PDiagnose~\cite{PDiagnose}) or via multi-task frameworks (Eadro~\cite{Eadro}, Dejavu~\cite{Dejavu}). Single-task methods miss shared diagnostic signals, while multi-task approaches often rely on labels or handcrafted rules. Self-supervised learning (SSL)~\cite{SSL} provides a scalable alternative: ART~\cite{ART} jointly supports AD, FT, and RCL without labels, and TrioXpert~\cite{TrioXpert} further improves interpretability via LLM-based reasoning.

\begin{figure}[h]
  \centering
  \includegraphics[width=0.6\linewidth]{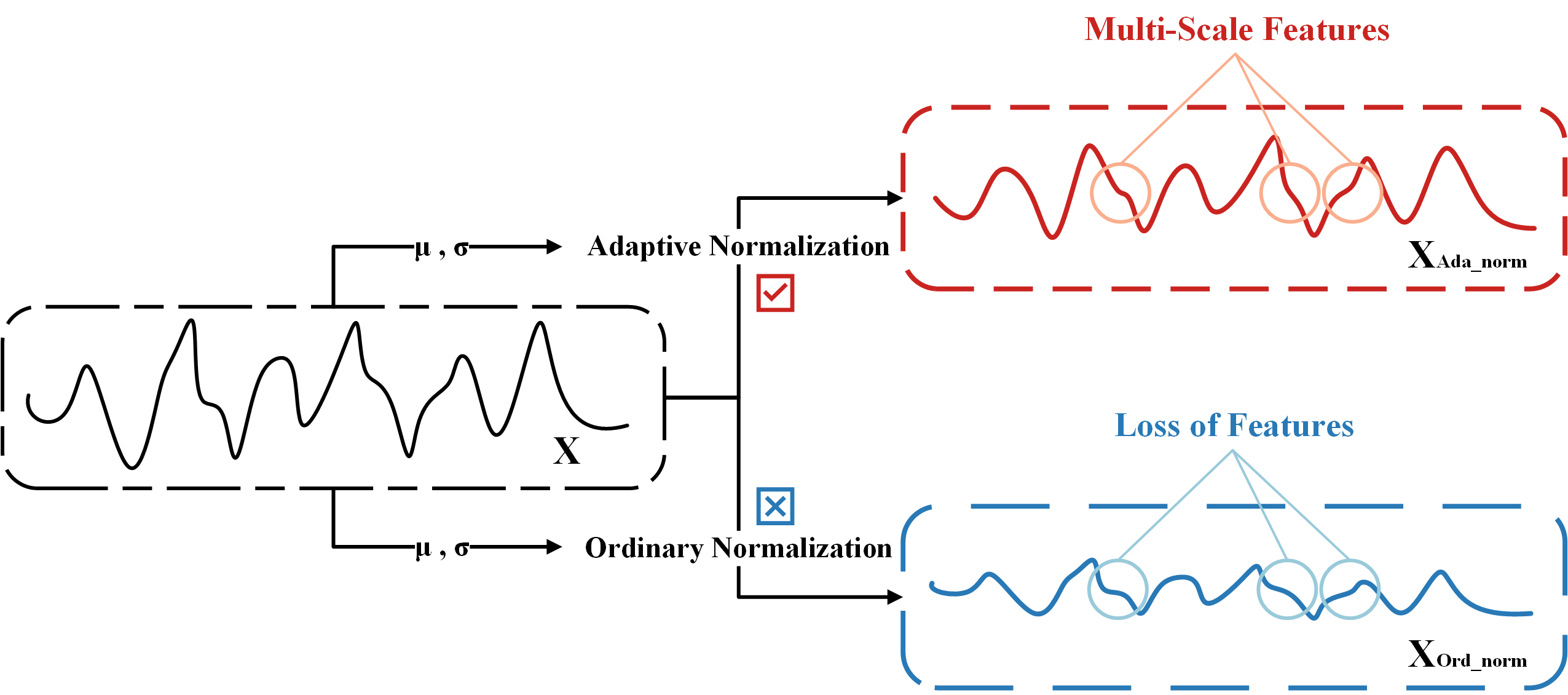}
  \caption{Static vs. adaptive normalization. Adaptive normalization preserves multi-scale temporal deviations, whereas static normalization smooths them out.}
  \label{fig:fig1}
\end{figure}

Despite methodological differences, most representation-centric methods use static normalization, misaligned with microservice data. Time series are non-stationary with multi-scale dynamics~\cite{TAFAS}, causing Z-score normalization~\cite{Z-score} to suppress informative deviations (Figure~\ref{fig:fig1}). Service call graphs are heterogeneous, yet graph encoders like GraphSAGE~\cite{GraphSAGE} apply uniform normalization, limiting expressiveness.

To address these challenges, we first formally define the problem of automated incident management in large-scale microservice systems. We consider a microservice system $\mathcal{S}$ has $N$ instances $\{s_1,\ldots,s_N\}$ linked by a directed call graph $G=(\mathcal{V},\mathcal{E})$, where $(s_i,s_j)\in\mathcal{E}$ indicates $s_i$ invokes $s_j$. Each instance emits \emph{metrics} $M_i\in\mathbb{R}^{T\times K_m}$, \emph{logs} $L_i\in\mathbb{R}^{T\times K_l}$, and \emph{traces} $R_i\in\mathbb{R}^{T\times K_r}$ over $T$ timestamps, forming a multimodal sequence $X_i^{(t)}\in\mathbb{R}^{T\times K}$ ($K=K_m+K_l+K_r$) and system state $X^{(t)}\in\mathbb{R}^{N\times T\times K}$. Given $X^{(t)}$, automated incident management jointly addresses: (i) AD: detect anomalies ($y^{(t)}\in\{0,1\}$); (ii) FT: classify into $C$ failure types $f\in\{1,\ldots,C\}$; (iii) RCL: rank all service instances $\{s_i\}$ by root-cause likelihood and return top-$k$ candidates. This is done without labeled data using only normal-period observations.

The fundamental challenges arise from (1) \emph{non-stationary temporal dynamics}: distribution shifts, concept drift, and transient spikes; and (2) \emph{structural heterogeneity}: directed, evolving call graphs with varying node degrees. Static normalization introduces \emph{representation bias}, degrading all tasks.

Based on this insight, we propose \textbf{STAR}, a spatial-temporal adaptive representation learning framework addressing temporal non-stationarity and structural heterogeneity via \emph{Temporal Adaptive Normalization (TAN)} and \emph{Spatial Adaptive Normalization (SAN)}. TAN normalizes multivariate time series with multi-scale temporal context, and SAN performs structure-aware normalization over service dependency graphs. Both integrate into a unified self-supervised framework supporting AD, FT, and RCL end-to-end. Code and data are available at https://github.com/XinhuaMiao/STAR.

The main contributions of this paper are as follows:
\begin{itemize}[leftmargin=1.5em]
\item We identify static normalization as a key source of representation bias and advocate adaptive, context-aware normalization for non-stationary, graph-structured data.
\item We propose STAR, a spatial-temporal adaptive representation learning framework jointly integrating TAN and SAN for multimodal observability data.
\item Extensive experiments on two real-world microservice benchmarks demonstrate that STAR outperforms all state-of-the-art baselines across all tasks.
\end{itemize}

\section{Related Work}
\subsection{Automated Incident Management}
Automated incident management in microservice systems involves anomaly detection (AD), failure triage (FT), and root cause localization (RCL). Early methods address these tasks separately, including anomaly detection~\cite{Hades,AnoTraceAE}, failure triage~\cite{MicroCBR}, and root cause localization~\cite{PDiagnose,Nezha}. Although effective in specific scenarios, they rely on task-specific pipelines and cannot exploit shared diagnostic signals.

Recent frameworks jointly model multiple diagnostic tasks~\cite{Dejavu,Eadro,DiagFusion}. LLM-based systems such as TrioXpert~\cite{TrioXpert} further improve interpretability through collaborative reasoning over multimodal data. However, most existing approaches depend on labels, handcrafted rules, or expensive reasoning components, limiting scalability in dynamic environments.

\begin{figure*}[t]
\centering
\includegraphics[width=0.65\linewidth]{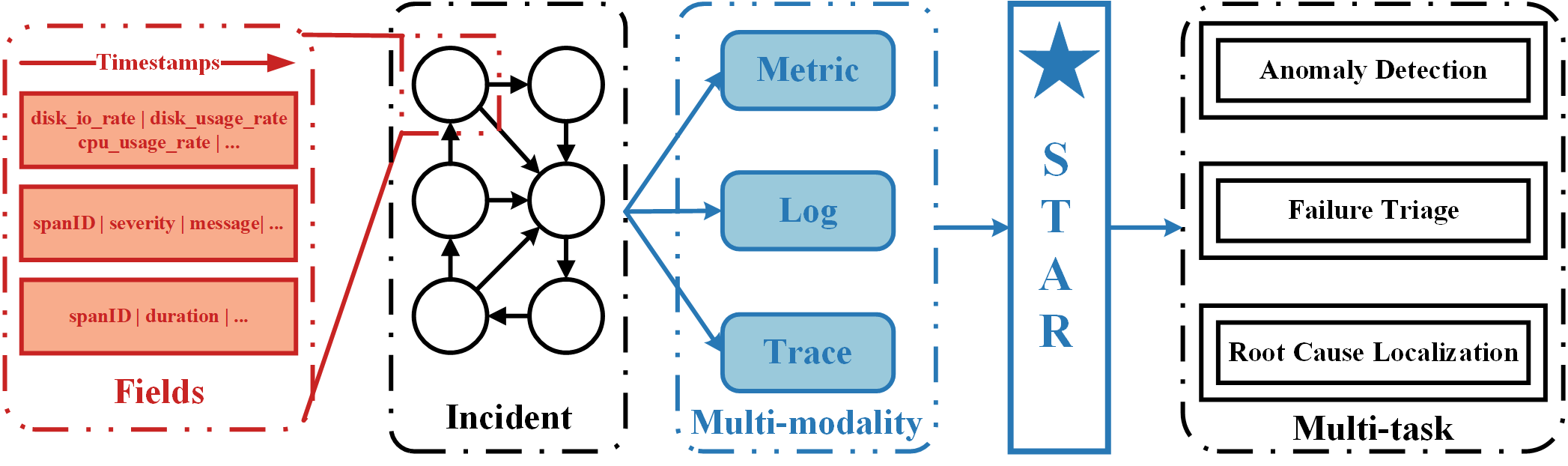}
\caption{Unified incident management frameworks encode multimodal observability data into shared representations for AD, FT, and RCL.}
\label{fig:fig2}
\end{figure*}

Figure~\ref{fig:fig2} illustrates the common paradigm of representation-centric incident management, where multimodal observability data are encoded into shared representations for multiple diagnostic tasks.

\subsection{Self-Supervised and Multimodal Representation Learning}

Self-supervised learning (SSL) enables representation learning without labels and has shown promise for unified incident management~\cite{SSL}. ART~\cite{ART} is the first SSL framework supporting AD, FT, and RCL jointly, using Transformers~\cite{Transformer}, recurrent networks~\cite{GRU1,GRU2}, and GraphSAGE~\cite{GraphSAGE} to model channel, temporal, and call dependencies.

Despite their effectiveness, existing SSL and multimodal frameworks primarily focus on feature fusion and dependency modeling, while largely overlooking representation bias caused by static normalization under non-stationary temporal dynamics and heterogeneous service structures.

\subsection{Adaptive Normalization}

Adaptive normalization conditions normalization parameters on input context to improve robustness under distribution shifts~\cite{BatchNorm,LayerNorm}. In time-series modeling, RevIN~\cite{RevIN} and Koopa~\cite{Koopa} address non-stationarity but rely on global statistics and target forecasting tasks. In graph learning, GraphNorm~\cite{GraphNorm} and DegNorm~\cite{DegNorm} adapt normalization to graph structure but do not explicitly model directed and heterogeneous service dependencies. To our knowledge, STAR is the first self-supervised framework that jointly integrates temporal and structural adaptive normalization for unified AD, FT, and RCL.

\section{Methodology}

\subsection{Overview}
\begin{figure*}[t]
  \centering
  \includegraphics[width=0.9\linewidth]{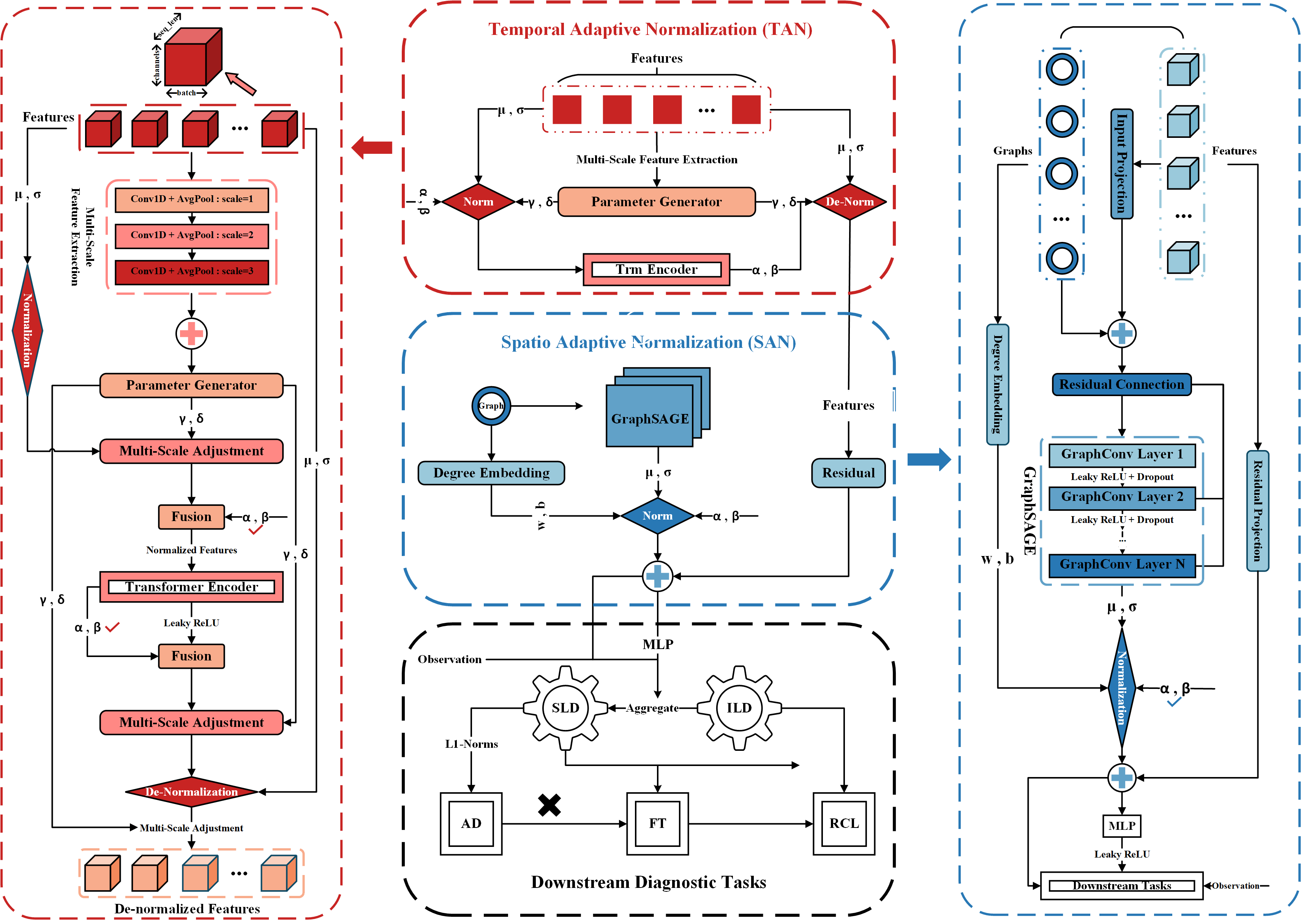}
  \caption{Overall STAR framework with two core components: TAN and SAN.}
  \label{fig:fig3}
\end{figure*}

As shown in Figure~\ref{fig:fig3}, STAR addresses the complexities of microservice systems via two adaptive normalization mechanisms: \emph{Temporal Adaptive Normalization (TAN)} and \emph{Spatial Adaptive Normalization (SAN)}. Together, they capture rich spatial-temporal dependencies, enabling robust, unsupervised AD, FT, and RCL.

\subsection{Temporal Adaptive Normalization (TAN)} \label{3.2}
Multimodal time series in microservices are non-stationary and channel-heterogeneous, which conventional normalization (Batch~\cite{BatchNorm}, Layer~\cite{LayerNorm}) cannot handle. TAN leverages multi-scale temporal context to dynamically generate normalization parameters, preserving informative patterns across time scales and improving robustness under non-stationary dynamics.

The forward computation of the multi-scale adaptive normalization consists of the following steps:

\subsubsection{Multi-Scale Contextual Feature Extraction}
Given an input tensor $X\in\mathbb{R}^{B\times L\times C}$, where $B$ is the batch size, $L$ is the sequence length, and $C$ is the number of channels, we first transpose it to $X^{\prime}\in\mathbb{R}^{B\times C\times L}$ to apply a one-dimensional convolution. Subsequently, we use a set of convolutional layers with different kernel sizes and global average pooling layers to extract multi-scale features:
\begin{equation}F_{\mathrm{scale}}^i=\text{AdaptiveAvgPool}1\mathrm{d}(\mathrm{Conv}1\mathrm{d}(X^{\prime},k^i)),\end{equation}
where $i=1,2,...,N_s$ and $k^i$ is the convolution kernel size corresponding to the $i^{th}$ scale, and $N_s$ is the total number of scales. Since it contains metric, log and trace, we set ${N_s}=3$. Each $F_{\mathrm{scale}}^i\in\mathbb{R}^{B\times C\times1}$ is compressed into a dimension $\mathbb{R}^{B\times C}$, and then the features of all scales are spliced on the last dimension to form a multi-scale context feature $F_{\mathrm{context}}\in\mathbb{R}^{B\times(N_s\cdot C)}$.
\subsubsection{Adaptive Parameter Generation}
The concatenated multi-scale context feature $F_{context}$ is input into a parameter generator composed of a fully connected layer to generate the parameters required for adaptive normalization:
\begin{equation}\label{eq:tan_params}[\gamma_{1},\delta_{1},\gamma_{2},\delta_{2}]=\text{ParameterGenerator}(F_{\mathrm{context}}),\end{equation}

  where the output is divided into four equal parts,
  $\gamma_1,\delta_1,\gamma_2,\delta_2\in\mathbb{R}^{B\times C}$, which serve as
   the scale and shift parameters in the subsequent adaptive normalization
  steps. Here we use $\delta$ (not $\beta$) to avoid confusion with the
  learnable fusion bias $\beta$ introduced in Eq.~(\ref{eq:fusion}).
\subsubsection{Adaptive Normalization}
\par First, we compute the mean and standard deviation of the input $X$ along the sequence dimension ($L$):
\begin{equation}\mu=\frac{1}{L}\sum_{l=1}^LX_{[:,l,:]},\quad\sigma=\sqrt{\frac{1}{L}\sum_{l=1}^L(X_{[:,l,:]}-\mu)^2+\epsilon},\end{equation}
where $\epsilon$ is a very small constant used to maintain numerical stability. Next, we perform a two-step normalization operation. First, perform basic normalization:
\begin{equation}\hat{X}_{\mathrm{base}}=\frac{X-\mu}{\sigma},\end{equation}
Then, the parameters generated in the first step are used for adaptive transformation:
\begin{equation}\hat{X}_{\mathrm{adapt}}=\hat{X}_{\mathrm{base}}\odot\gamma_{1}+\delta_{1},\end{equation}

Finally, we use the learnable weight parameters $\alpha\in\mathbb{R}^{1\times1\times C}$ and $\beta\in\mathbb{R}^{1\times1\times C}$ to perform weighted fusion on the basic normalization result and the adaptive transformation result to obtain the final normalized output:
\begin{equation}\label{eq:fusion}\hat{X}=\alpha\odot\hat{X}_\mathrm{base}+(1-\alpha)\odot\hat{X}_\mathrm{adapt}+\beta,\end{equation}
All parameters in this step $(\mu,\sigma,\gamma_1,\beta_1,\gamma_2,\beta_2)$ will be saved for possible subsequent de-normalization operations.
\subsubsection{De-Normalization}
After the Transformer encoder of the model, we need to restore the features to near the original scale. Therefore, temporal adaptive normalization (TAN) provides a corresponding de-normalization process:
\begin{equation}X_{\mathrm{recon}}=\left(\frac{(\hat{X}-\beta)/\alpha-\delta_1}{\gamma_1+\epsilon}\right)\odot\sigma+\mu,\end{equation}
Then, further transformation is performed using the second set of parameters:
\begin{equation}X_{\mathrm{final}}=X_{\mathrm{recon}}\odot\gamma_2+\delta_2.\end{equation}
where we get the final output $X_{final}$ of the temporal part.

Through the above design, TAN dynamically conditions normalization on multi-scale temporal context, improving the representation of heterogeneous multimodal time series. We integrate TAN into the Transformer encoder to enhance channel-wise dependency (CHA) modeling.
\subsection{Spatial Adaptive Normalization (SAN)}
In this section, we will introduce the innovative graph structure adaptive normalization that is coordinated by structure-awareness and residual connections.
\subsubsection{Graph structure adaptive normalization} \label{3.3.1}
In graph neural networks (GNNs) \cite{GNN}, conventional normalization applies uniform scaling across nodes, ignoring structural heterogeneity. We address this by introducing \emph{Graph Adaptive Normalization}, which conditions normalization parameters on local structural properties (e.g., node degree), enabling graph-aware normalization and more effective representation learning over service dependency graphs.

The forward computation of graph adaptive normlization is as follows: 
\paragraph{Node Degree Feature Extraction.}
For a given graph $G$, we first obtain the in-degree $d_i$ of each node i and clip it to a preset reasonable range to prevent extreme values:
\begin{equation}\mathrm{degrees}=\mathrm{clip}(G.\mathrm{in_{degrees}}(),0,D-1),\end{equation}
where $D$ is the maximum degree we set.
We then map the discrete degrees into a continuous feature vector through a learnable embedding layer:
\begin{equation}e_i=\text{DegreeEmbedding}(d_i),\end{equation}
where $e_i\in\mathbb{R}^C$ and $C$ is the dimension of node features.

This embedding captures the potential common behavior patterns of nodes with different degrees.
\paragraph{Structural Adaptive Parameter Generation.}
We compute the graph-level degree-dependent scaling and bias components based on the degree embeddings of all nodes. Specifically, we compute the mean of all node embeddings as the adaptive scaling weight $w_{deg}$, and the standard deviation as the adaptive bias component $b_{deg}$:
\begin{equation}\label{eq:san_degree}w_{\mathrm{deg}}=\frac{1}{N}\sum_{i=1}^Ne_i,\quad
\quad{b}_{\mathrm{deg}}=\sqrt{\frac{1}{N}\sum_{i=1}^N({e}_i-{w}_{\mathrm{deg}})^2},\end{equation}
where ${w}_{\mathrm{deg}},{b}_{\mathrm{deg}}\in\mathbb{R}^{1\times C}$ is a graph-level statistic that provides a structural prior for the normalization process of the entire graph.
\paragraph{Graph-Structure-Aware Normalization.}
For the input node features ${X}\in\mathbb{R}^{N\times C}$, we first calculate the standardized mean and standard deviation along the node dimension ($N$):
\begin{equation}{\mu}=\frac{1}{N}\sum_{i=1}^{N}{x}_{i},\quad
\quad{\sigma}=\sqrt{\frac{1}{N}\sum_{i=1}^{N}({x}_{i}-{\mu})^{2}+\epsilon},\end{equation}
where $\epsilon$ is a very small constant used to maintain numerical stability. Next, we perform a preliminary normalization:
\begin{equation}\hat{{X}}_{\mathrm{standard}}=\frac{{X}-{\mu}}{{\sigma}},\end{equation}
The most critical step is that we combine the standard learnable parameters ${\alpha}_\mathrm{base},{\beta}_\mathrm{base}\in\mathbb{R}^{1\times C}$ with the adaptive parameters ${w}_{\mathrm{deg}},{b}_{\mathrm{deg}}$ derived from the graph structure to generate the final structure-aware normalization parameters:
\begin{equation}{w}_{\mathrm{final}}={\alpha}_{\mathrm{base}}+{w}_{\mathrm{deg}},\quad
\quad{b}_{\mathrm{final}}={\beta}_{\mathrm{base}}+{b}_{\mathrm{deg}},\end{equation}
Finally, the fused parameters are used to scale and offset the normalized node features:
\begin{equation}\hat{{X}}=\hat{{X}}_{\mathrm{standard}}\odot{w}_{\mathrm{final}}+{b}_{\mathrm{final}},\end{equation}
where $\odot$ represents element-by-element multiplication, which is the broadcast mechanism.
\paragraph{Integrated on GraphSAGE Encoder.}
We integrate the graph adaptive normlization layer into the end of the improved GraphSAGE encoder. This encoder also introduces residual connections to ensure training stability and alleviate the vanishing gradient problem. Its forward propagation process can be briefly described as:
\begin{gather}\label{eq:san_residual}
\begin{aligned}
h^{(0)} 
&= \mathrm{InputProj}(X), \\
h^{(l)} 
&= \mathrm{GraphConv}^{(l)}(h^{(l-1)}) + h^{(l-1)}, \ \text{(for residual layers)}, \\
h^{(\mathrm{out})} 
&= \mathrm{GraphAdaptiveNorm}(h^{(L)}).
\end{aligned}
\end{gather}
In this way, GraphSAGE encoder not only aggregates neighbor information, but its output features are also adaptively calibrated according to the global graph structure.
\subsubsection{Structure-awareness and residual learning}
While GraphSAGE aggregates dependencies via stacked graph convolutions, deep architectures can suffer from representation bottlenecks and optimization instability. We address this by incorporating flexible dimension projection, residual connections, and structure-adaptive normalization, yielding more stable and expressive node representation learning.

\paragraph{Input Projection.}
In the original implementation, input features are fed directly into the graph convolution layer. This can cause information bottlenecks or dimensionality waste when the input dimension does not match the hidden layer dimension. To address this, we add a learnable linear projection layer to ensure that features are mapped to a unified latent space before entering the graph convolution:
\begin{equation}
\begin{aligned}
h^{(0)} 
&= \mathrm{InputProj}(X) =
\begin{cases}
W_{\mathrm{proj}} X + b_{\mathrm{proj}}, 
& d_{\mathrm{in}} \neq d_{\mathrm{hidden}}, \\
X, 
& d_{\mathrm{in}} = d_{\mathrm{hidden}},
\end{cases}
\end{aligned}
\end{equation}

where ${W}_{\mathrm{proj}}\in\mathbb{R}^{d_{\mathrm{hidden}}\times d_{\mathrm{in}}},{b}_{\mathrm{proj}}\in\mathbb{R}^{d_{\mathrm{hidden}}}$ are projection parameters. 
\par This operation enhances the model's flexibility for input data of different dimensions.
\paragraph{Residual Graph Convolutional Layers.}
The training of deep graph neural networks is often plagued by the vanishing gradient problem. Inspired by ResNet \cite{ResNet}, we introduce residual connections in all graph convolutional layers except the first to facilitate direct backpropagation of gradients while preserving underlying feature information.
\par Let $\mathcal{G}^{(l)}$ be the graph convolution operation in the $l^{th}$ layer, then the forward propagation of this layer is defined as:
\begin{equation}\begin{aligned}
& h_{res}={h}^{(l-1)} ,\quad{h}^{(l)}=\mathcal{G}^{(l)}({h}^{(l-1)}) ,\quad{h}^{(l)}={h}^{(l)}+{h}_{\mathrm{res}} \ \text{(Add Residual),}
\end{aligned}\end{equation}
where $l=1,...,L$, and in the actual code implementation, we skip the residual connection of the first layer by judging $i>0$ and perform the addition operation only after ensuring the dimension matching, thus ensuring the stability of the calculation.
\paragraph{Integration of Graph Adaptive Normalization.}
The original normalization parameters are globally shared and independent of the graph structure. We integrate the graph adaptive normalization layer proposed in section \ref{3.3.1} into the end of the encoder to perform structure-adaptive calibration on the final output node representation. This operation can be formally expressed as:
\begin{equation}\hat{{H}}=\text{GraphAdaptive}\mathrm{Norm}({h}^{(L)},G),\end{equation}
where ${H}^{(L)}$ is the output feature after $L$ layers of graph convolution, and $G$ is the input graph. 
\par This normalization layer uses the in-degree information of nodes to dynamically generate scaling and offset parameters, making the normalization process aware of the importance of nodes in the graph and thus learning more discriminative representations.
\paragraph{Final Output with Skip Connection.}
To ensure that the decoder or downstream tasks can fully utilize the information extracted from the original input, we add an additional skip connection from the input directly to the output. This connection projects the dimension of the input feature ${X}$ to the same output dimension through a projection layer similar to the input projection above:
\begin{equation}{h}_{\mathrm{out}}=\hat{{H}}+\mathrm{Proj}({X}).\end{equation}
At this point, the complete forward propagation process of the graph encoder is completed and the final output is obtained.

Together, these refinements enhance the representation capacity of the GraphSAGE encoder by accommodating heterogeneous node features, stabilizing deep propagation, and injecting structural information into node representations, leading to more effective modeling of graph dependencies.

\section{Experiments}
\subsection{Experimental setup}
\subsubsection{Datasets}

We conduct experiments using the same two datasets, D1 and D2, as the original ART dataset to ensure comparability, which are shown in Table \ref{tab:table1}. D1 is from a simulated e-commerce system and contains 46 instances; D2 is from a bank management system and contains 18 instances. Both datasets contain metrics, logs, and traces. The split between the training and test sets follows the original setup. For more details about the datasets, please refer to Appendix \ref{appendix:datasets}.

\begin{table*}[t]
\caption{The details of the D1 and D2 datasets.}
\centering
\label{tab:table1}
\begin{tabular}{l|l|l|l|l|l|l|l}
\hline
Datasets & Instances & Normal & Failures & Failure Types & Metrics & Logs   & Traces \\ \hline
D1       & 46        & 3,714  & 210      & 5             & 20.92M  & 66.65M & 44.86M \\ \hline
D2       & 18        & 12,297 & 133      & 6             & 12.87M  & 21.36M & 21.43M \\ \hline
\end{tabular}
\end{table*}

\subsubsection{Baselines}
To ensure the scientific nature of the comparative experiments, we use the same benchmark models as the TrioXpert method~\cite{TrioXpert} for comparison:\\
\textit{Single-task.} Hades~\cite{Hades} (AD), MicroCBR~\cite{MicroCBR} (FT), and PDiagnose~\cite{PDiagnose} (RCL).\\
\textit{Multi-task.} TrioXpert~\cite{TrioXpert}, ART~\cite{ART}, Eadro~\cite{Eadro}, Dejavu~\cite{Dejavu}, and DiagFusion~\cite{DiagFusion}.

\subsubsection{Evaluation Metrics}
AD is a binary classification problem of whether a fault occurs, while FT is a multi-classification problem of the type of the current fault. Therefore, for both AD and FT tasks, we use precision, recall, and f1-score as evaluation metrics. For the RCL task, we introduce the top-k accuracy $Top-k=\frac{1}{N}\sum_{i=1}^N\mathbb{I}\left(g_i\in\mathrm{Top}_k(P_i)\right)$ to calculate the probability of the root cause among the first $k$ predicted candidates, where $N$ is the number of evaluation failures, $g_i$ is the ground truth root cause of the $i^{th}$ failure case, and $\mathrm{Top}_k(P_i)$ is the set of top-k root causes predicted by the model for the $i^{th}$ failure case. In addition, we will also calculate the average accuracy of the top-k accuracy (k value ranges from 1 to 5). For the implementation details, please refer to Appendix \ref{appendix:i}.

\begin{table*}[t]
\caption{Comparative experimental results of the AD, FT, and RCL tasks between the STAR method and the baseline models. The best performance is highlighted in \textbf{bold} and the second best is \underline{underlined}. $M_s$ represents the \textit{Single-task} method.}
\centering
\label{tab:table3}
\scalebox{0.88}{\begin{tabular}{l|l|lll|lll|lll}
\hline
\multirow{2}{*}{Datasets} & \multirow{2}{*}{Methods} & \multicolumn{3}{l|}{AD}                                                                    & \multicolumn{3}{l|}{FT}                                                                    & \multicolumn{3}{l}{RCL}                                                                    \\ \cline{3-11} 
                          &                          & \multicolumn{1}{l|}{Precision}      & \multicolumn{1}{l|}{Recall}         & F1-score       & \multicolumn{1}{l|}{Precision}      & \multicolumn{1}{l|}{Recall}         & F1-score       & \multicolumn{1}{l|}{Top-1}          & \multicolumn{1}{l|}{Top-3}          & AVG@5          \\ \hline
\multirow{7}{*}{D1}       & $M_s$                       & \multicolumn{1}{l|}{0.866}          & \multicolumn{1}{l|}{0.863}          & 0.865          & \multicolumn{1}{l|}{0.667}          & \multicolumn{1}{l|}{0.796}          & 0.717          & \multicolumn{1}{l|}{0.615}          & \multicolumn{1}{l|}{0.692}          & 0.685          \\
                          & DiagFusion               & \multicolumn{1}{l|}{-}              & \multicolumn{1}{l|}{-}              & -              & \multicolumn{1}{l|}{0.675}          & \multicolumn{1}{l|}{0.500}          & 0.568          & \multicolumn{1}{l|}{0.310}          & \multicolumn{1}{l|}{0.452}          & 0.467          \\
                          & Dejevu                   & \multicolumn{1}{l|}{-}              & \multicolumn{1}{l|}{-}              & -              & \multicolumn{1}{l|}{0.369}          & \multicolumn{1}{l|}{0.621}          & 0.415          & \multicolumn{1}{l|}{0.411}          & \multicolumn{1}{l|}{0.679}          & 0.625          \\
                          & Eadro                    & \multicolumn{1}{l|}{0.425}          & \multicolumn{1}{l|}{0.946}          & 0.586          & \multicolumn{1}{l|}{-}              & \multicolumn{1}{l|}{-}              & -              & \multicolumn{1}{l|}{0.137}          & \multicolumn{1}{l|}{0.315}          & 0.302          \\
                          & ART                      & \multicolumn{1}{l|}{{\ul 0.899}}    & \multicolumn{1}{l|}{{\ul 0.990}}    & {\ul 0.942}    & \multicolumn{1}{l|}{0.836}          & \multicolumn{1}{l|}{{\ul 0.809}}    & {\ul 0.812}    & \multicolumn{1}{l|}{{\ul 0.667}}    & \multicolumn{1}{l|}{\textbf{0.810}} & {\ul 0.776}    \\
                          & TrioXpert                & \multicolumn{1}{l|}{0.880}          & \multicolumn{1}{l|}{0.972}          & 0.924          & \multicolumn{1}{l|}{{\ul 0.852}}    & \multicolumn{1}{l|}{0.768}          & 0.807          & \multicolumn{1}{l|}{0.651}          & \multicolumn{1}{l|}{0.778}          & 0.773          \\
                          & STAR                     & \multicolumn{1}{l|}{\textbf{0.916}} & \multicolumn{1}{l|}{\textbf{1.000}} & \textbf{0.956} & \multicolumn{1}{l|}{\textbf{0.864}} & \multicolumn{1}{l|}{\textbf{0.881}} & \textbf{0.870} & \multicolumn{1}{l|}{\textbf{0.679}} & \multicolumn{1}{l|}{{\ul 0.798}}    & \textbf{0.793} \\ \hline
\multirow{7}{*}{D2}       & $M_s$                       & \multicolumn{1}{l|}{0.867}          & \multicolumn{1}{l|}{0.868}          & 0.868          & \multicolumn{1}{l|}{0.629}          & \multicolumn{1}{l|}{0.678}          & 0.636          & \multicolumn{1}{l|}{0.037}          & \multicolumn{1}{l|}{0.296}          & 0.285          \\
                          & DiagFusion               & \multicolumn{1}{l|}{-}              & \multicolumn{1}{l|}{-}              & -              & \multicolumn{1}{l|}{0.797}          & \multicolumn{1}{l|}{0.527}          & 0.593          & \multicolumn{1}{l|}{0.582}          & \multicolumn{1}{l|}{0.709}          & 0.695          \\
                          & Dejevu                   & \multicolumn{1}{l|}{-}              & \multicolumn{1}{l|}{-}              & -              & \multicolumn{1}{l|}{0.718}          & \multicolumn{1}{l|}{0.340}          & 0.417          & \multicolumn{1}{l|}{0.402}          & \multicolumn{1}{l|}{0.667}          & 0.619          \\
                          & Eadro                    & \multicolumn{1}{l|}{0.767}          & \multicolumn{1}{l|}{0.935}          & 0.842          & \multicolumn{1}{l|}{-}              & \multicolumn{1}{l|}{-}              & -              & \multicolumn{1}{l|}{0.157}          & \multicolumn{1}{l|}{0.315}          & 0.310          \\
                          & ART                      & \multicolumn{1}{l|}{{\ul 0.877}}    & \multicolumn{1}{l|}{{\ul 0.960}}    & {\ul 0.917}    & \multicolumn{1}{l|}{{\ul 0.851}}    & \multicolumn{1}{l|}{{\ul 0.796}}    & {\ul 0.802}    & \multicolumn{1}{l|}{{\ul 0.722}}    & \multicolumn{1}{l|}{{\ul 0.889}}    & {\ul 0.870}    \\
                          & TrioXpert                & \multicolumn{1}{l|}{0.854}          & \multicolumn{1}{l|}{0.972}          & 0.909          & \multicolumn{1}{l|}{0.814}          & \multicolumn{1}{l|}{0.725}          & 0.767          & \multicolumn{1}{l|}{0.550}          & \multicolumn{1}{l|}{0.775}          & 0.750          \\
                          & STAR                     & \multicolumn{1}{l|}{\textbf{0.974}} & \multicolumn{1}{l|}{\textbf{1.000}} & \textbf{0.987} & \multicolumn{1}{l|}{\textbf{0.876}} & \multicolumn{1}{l|}{\textbf{0.852}} & \textbf{0.855} & \multicolumn{1}{l|}{\textbf{0.741}} & \multicolumn{1}{l|}{\textbf{0.926}} & \textbf{0.885} \\ \hline
\end{tabular}}
\end{table*}

\subsection{Overall performance}
In this section, we present the experimental results of STAR compared with other baseline models on AD, FT and RCL tasks as shown in Table \ref{tab:table3}. STAR significantly outperforms all baseline methods on the AD task, achieving an F1-score of 0.956 on D1 and 0.983 on D2, highlighting the effectiveness of multi-scale adaptive normalization in improving the representation of multimodal time series from a temporal perspective and capturing abnormal signals. On the FT task, STAR achieves a weighted F1 score of 0.870 on D1 and 0.855 on D2. This demonstrates that graph structure adaptive normalization, by introducing node degree-aware normalization from a spatial perspective, enhances the discriminability of graph structural features and improves classification accuracy. For the RCL task, STAR achieves state-of-the-art (SOTA) performance, with an AVG@5 of 0.793 on D1 and 0.885 on D2, excluding the top-3 accuracy on D1.

Overall, STAR significantly outperforms all baseline methods on all three tasks, demonstrating the effectiveness of its unified representation learning and adaptive normalization mechanism.

\subsection{Comparison with Alternative Normalization Methods.}
To further validate the design choices in TAN and SAN, we replace each
component with representative normalization methods from the literature and
re-evaluate on both datasets.
For the temporal domain, we compare TAN against plain Z-score normalization (the ART baseline \cite{ART}), standard BatchNorm \cite{BatchNorm}, LayerNorm \cite{LayerNorm}, and RevIN~\cite{RevIN}. For the spatial domain, we compare SAN against GraphNorm~\cite{GraphNorm}, node-level BatchNorm \cite{BatchNorm}, and DegNorm \cite{DegNorm}. Table~\ref{tab:norm_comparison} reports the results.

\begin{table*}[h]
  \centering
  \caption{Comparison with alternative normalization methods on D1 and D2.
  \textbf{Bold} indicates the best result per column. TAN and SAN (ours) achieve
   the best or near-best performance across all settings, confirming the
  effectiveness of context-conditioned adaptive normalization over static
  alternatives.}
  \label{tab:norm_comparison}
\scalebox{0.93}{\begin{tabular}{l|lll|lll|lll}
\hline
\multirow{2}{*}{Methods} & \multicolumn{3}{l|}{AD}                                                                    & \multicolumn{3}{l|}{FT}                                                                    & \multicolumn{3}{l}{RCL}                                                                    \\ \cline{2-10} 
                         & \multicolumn{1}{l|}{Precision}      & \multicolumn{1}{l|}{Recall}         & F1-score       & \multicolumn{1}{l|}{Precision}      & \multicolumn{1}{l|}{Recall}         & F1-score       & \multicolumn{1}{l|}{Top-1}          & \multicolumn{1}{l|}{Top-3}          & AVG@5          \\ \hline
ZScore                   & \multicolumn{1}{l|}{0.884}          & \multicolumn{1}{l|}{0.929}          & 0.906          & \multicolumn{1}{l|}{0.755}          & \multicolumn{1}{l|}{0.750}          & 0.745          & \multicolumn{1}{l|}{{\ul 0.643}}    & \multicolumn{1}{l|}{{\ul 0.774}}    & {\ul 0.757}    \\
BatchNorm                & \multicolumn{1}{l|}{0.888}          & \multicolumn{1}{l|}{0.969}          & 0.927          & \multicolumn{1}{l|}{{\ul 0.792}}    & \multicolumn{1}{l|}{{\ul 0.798}}    & {\ul 0.792}    & \multicolumn{1}{l|}{0.619}          & \multicolumn{1}{l|}{0.702}          & 0.721          \\
LayerNorm                & \multicolumn{1}{l|}{0.856}          & \multicolumn{1}{l|}{0.908}          & 0.881          & \multicolumn{1}{l|}{0.737}          & \multicolumn{1}{l|}{0.702}          & 0.704          & \multicolumn{1}{l|}{0.607}          & \multicolumn{1}{l|}{0.762}          & 0.736          \\
RevIN                    & \multicolumn{1}{l|}{{\ul 0.890}}    & \multicolumn{1}{l|}{{\ul 0.990}}    & {\ul 0.937}    & \multicolumn{1}{l|}{0.762}          & \multicolumn{1}{l|}{0.762}          & 0.758          & \multicolumn{1}{l|}{0.631}          & \multicolumn{1}{l|}{0.750}          & 0.733          \\
\textbf{TAN}             & \multicolumn{1}{l|}{\textbf{0.899}} & \multicolumn{1}{l|}{\textbf{1.000}} & \textbf{0.947} & \multicolumn{1}{l|}{\textbf{0.820}} & \multicolumn{1}{l|}{\textbf{0.821}} & \textbf{0.818} & \multicolumn{1}{l|}{\textbf{0.655}} & \multicolumn{1}{l|}{\textbf{0.798}} & \textbf{0.779} \\ \hline
GraphNorm                & \multicolumn{1}{l|}{0.877}          & \multicolumn{1}{l|}{0.949}          & 0.912          & \multicolumn{1}{l|}{0.755}          & \multicolumn{1}{l|}{0.762}          & 0.755          & \multicolumn{1}{l|}{0.595}          & \multicolumn{1}{l|}{0.702}          & 0.705          \\
BatchNorm-G              & \multicolumn{1}{l|}{\textbf{0.917}} & \multicolumn{1}{l|}{0.898}          & 0.907          & \multicolumn{1}{l|}{{\ul 0.798}}    & \multicolumn{1}{l|}{0.786}          & 0.783          & \multicolumn{1}{l|}{0.619}          & \multicolumn{1}{l|}{0.691}          & 0.710          \\
DegNorm                  & \multicolumn{1}{l|}{0.889}          & \multicolumn{1}{l|}{{\ul 0.980}}    & {\ul 0.932}    & \multicolumn{1}{l|}{0.792}          & \multicolumn{1}{l|}{{\ul 0.798}}    & {\ul 0.787}    & \multicolumn{1}{l|}{{\ul 0.631}}    & \multicolumn{1}{l|}{{\ul 0.750}}    & {\ul 0.743}    \\
\textbf{SAN}             & \multicolumn{1}{l|}{{\ul 0.907}}    & \multicolumn{1}{l|}{\textbf{0.990}} & \textbf{0.946} & \multicolumn{1}{l|}{\textbf{0.822}} & \multicolumn{1}{l|}{\textbf{0.821}} & \textbf{0.819} & \multicolumn{1}{l|}{\textbf{0.655}} & \multicolumn{1}{l|}{\textbf{0.821}} & \textbf{0.786} \\ \hline
\end{tabular}}
\end{table*}

TAN outperforms all temporal baselines because it jointly learns
instance-specific scale and shift via multi-scale temporal context, rather
than applying fixed statistics or globally shared affine parameters.
SAN outperforms GraphNorm and DegNorm because it conditions the normalization
parameters on degree embeddings, enabling each node to receive structure-aware
adjustment rather than a uniform graph-level shift.
These results confirm that the design of adaptive, context-dependent
normalization is both necessary and sufficient for the observed performance
gains.

\subsection{Ablation Study}
To validate the effectiveness of the proposed modules, temporal adaptive normalization (TAN) and spatial adaptive normalization (SAN), we design the following ablation variants for comparison with STAR to verify the contributions of TAN and SAN: \textbf{Baseline.} Removes both TAN and SAN, i.e., the original ART method. \textbf{w/o TAN.} Removes only TAN. \textbf{w/o SAN.} Removes only SAN.

\begin{table*}[t]
\caption{The experiment results of ablation study on AD, FT, and RCL tasks. The best results are highlighted in bold.}
\centering
\label{tab:table6}
\scalebox{0.88}{\begin{tabular}{l|l|lll|lll|lll}
\hline
\multirow{2}{*}{Datasets} & \multirow{2}{*}{Methods} & \multicolumn{3}{l|}{AD}                                                                    & \multicolumn{3}{l|}{FT}                                                                    & \multicolumn{3}{l}{RCL}                                                                    \\ \cline{3-11} 
                          &                          & \multicolumn{1}{l|}{Precision}      & \multicolumn{1}{l|}{Recall}         & F1-score       & \multicolumn{1}{l|}{Precision}      & \multicolumn{1}{l|}{Recall}         & F1-score       & \multicolumn{1}{l|}{Top-1}          & \multicolumn{1}{l|}{Top-3}          & AVG@5          \\ \hline
\multirow{4}{*}{D1}       & Baseline                 & \multicolumn{1}{l|}{0.899}          & \multicolumn{1}{l|}{0.990}          & 0.942          & \multicolumn{1}{l|}{0.836}          & \multicolumn{1}{l|}{0.809}          & 0.812          & \multicolumn{1}{l|}{0.667}          & \multicolumn{1}{l|}{0.810}          & 0.776          \\
                          & w/o TAN                  & \multicolumn{1}{l|}{0.907}          & \multicolumn{1}{l|}{0.990}          & 0.946          & \multicolumn{1}{l|}{0.822}          & \multicolumn{1}{l|}{0.821}          & 0.819          & \multicolumn{1}{l|}{0.655}          & \multicolumn{1}{l|}{\textbf{0.821}} & 0.786          \\
                          & w/o SAN                  & \multicolumn{1}{l|}{0.899}          & \multicolumn{1}{l|}{\textbf{1.000}}          & 0.947          & \multicolumn{1}{l|}{0.820}          & \multicolumn{1}{l|}{0.821}          & 0.818          & \multicolumn{1}{l|}{0.655}          & \multicolumn{1}{l|}{0.798}          & 0.779          \\
                          & STAR                     & \multicolumn{1}{l|}{\textbf{0.916}} & \multicolumn{1}{l|}{\textbf{1.000}} & \textbf{0.956} & \multicolumn{1}{l|}{\textbf{0.864}} & \multicolumn{1}{l|}{\textbf{0.881}} & \textbf{0.870} & \multicolumn{1}{l|}{\textbf{0.679}} & \multicolumn{1}{l|}{0.798}          & \textbf{0.793} \\ \hline
\multirow{4}{*}{D2}       & Baseline                 & \multicolumn{1}{l|}{0.877}          & \multicolumn{1}{l|}{0.960}          & 0.917          & \multicolumn{1}{l|}{0.851}          & \multicolumn{1}{l|}{0.796}          & 0.802          & \multicolumn{1}{l|}{0.722}          & \multicolumn{1}{l|}{0.889}          & 0.870          \\
                          & w/o TAN                  & \multicolumn{1}{l|}{0.943}          & \multicolumn{1}{l|}{0.993}          & 0.967          & \multicolumn{1}{l|}{0.859}          & \multicolumn{1}{l|}{0.796}          & 0.810          & \multicolumn{1}{l|}{0.722}          & \multicolumn{1}{l|}{0.907}          & 0.878          \\
                          & w/o SAN                  & \multicolumn{1}{l|}{0.894}          & \multicolumn{1}{l|}{0.966}          & 0.929          & \multicolumn{1}{l|}{0.852}          & \multicolumn{1}{l|}{0.796}          & 0.811          & \multicolumn{1}{l|}{\textbf{0.741}} & \multicolumn{1}{l|}{0.907}          & 0.874          \\
                          & STAR                     & \multicolumn{1}{l|}{\textbf{0.974}} & \multicolumn{1}{l|}{\textbf{1.000}} & \textbf{0.987} & \multicolumn{1}{l|}{\textbf{0.876}} & \multicolumn{1}{l|}{\textbf{0.852}} & \textbf{0.855} & \multicolumn{1}{l|}{\textbf{0.741}} & \multicolumn{1}{l|}{\textbf{0.926}} & \textbf{0.885} \\ \hline
\end{tabular}}
\end{table*}

We present the ablation results for the AD, FT, and RCL tasks in Table \ref{tab:table6}. The results show that STAR outperforms all ablation variants in almost all cases, and the variants w/o TAN and w/o SAN also surpass the baseline model, fully demonstrating the effectiveness and robustness of the proposed spatial-temporal adaptive normalizations. A complementary t-SNE visualization of the representation space before and after TAN and SAN is provided in Appendix~\ref{appendix:tsne}.

\subsection{Hyperparameter Analysis}
\begin{figure*}[t]
\centering
\subfigure[$N_s$ on AD task]{
\begin{minipage}[t]{0.45\textwidth}
\includegraphics[width=1\linewidth]{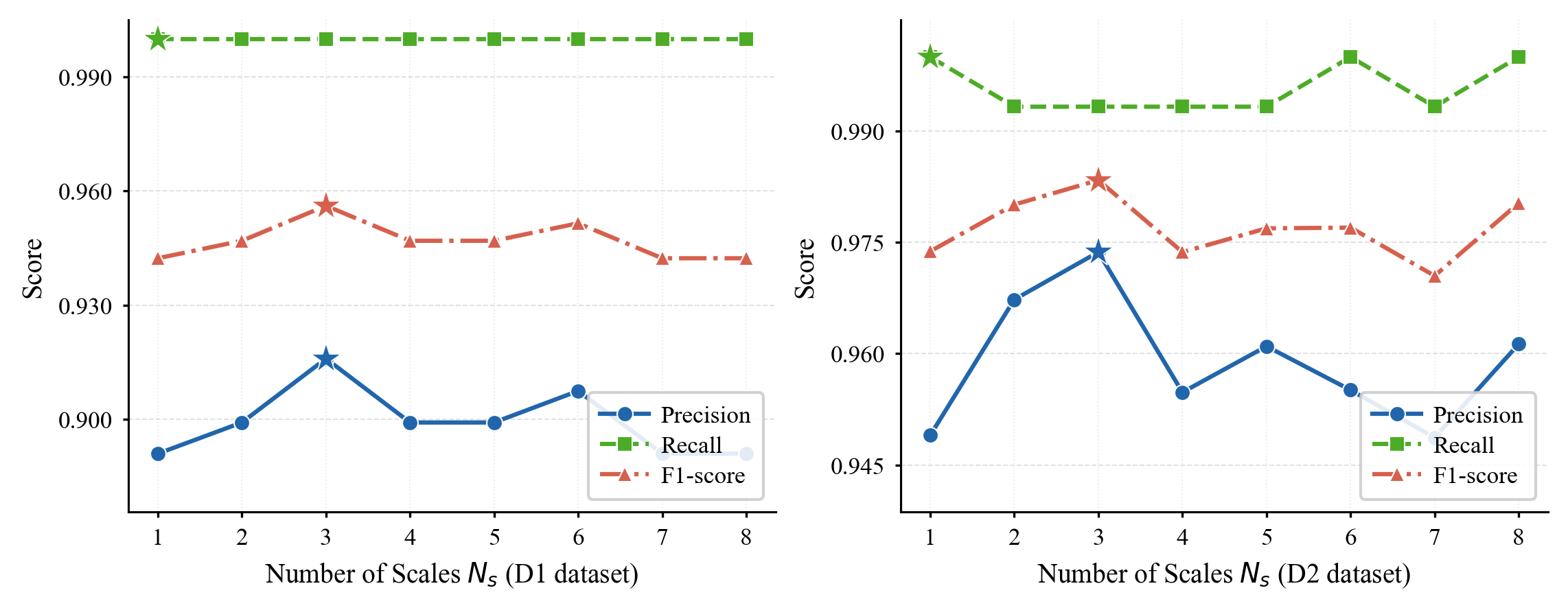}
\end{minipage}}
\subfigure[$D$ on AD task]{
\begin{minipage}[t]{0.45\textwidth}
\includegraphics[width=1\linewidth]{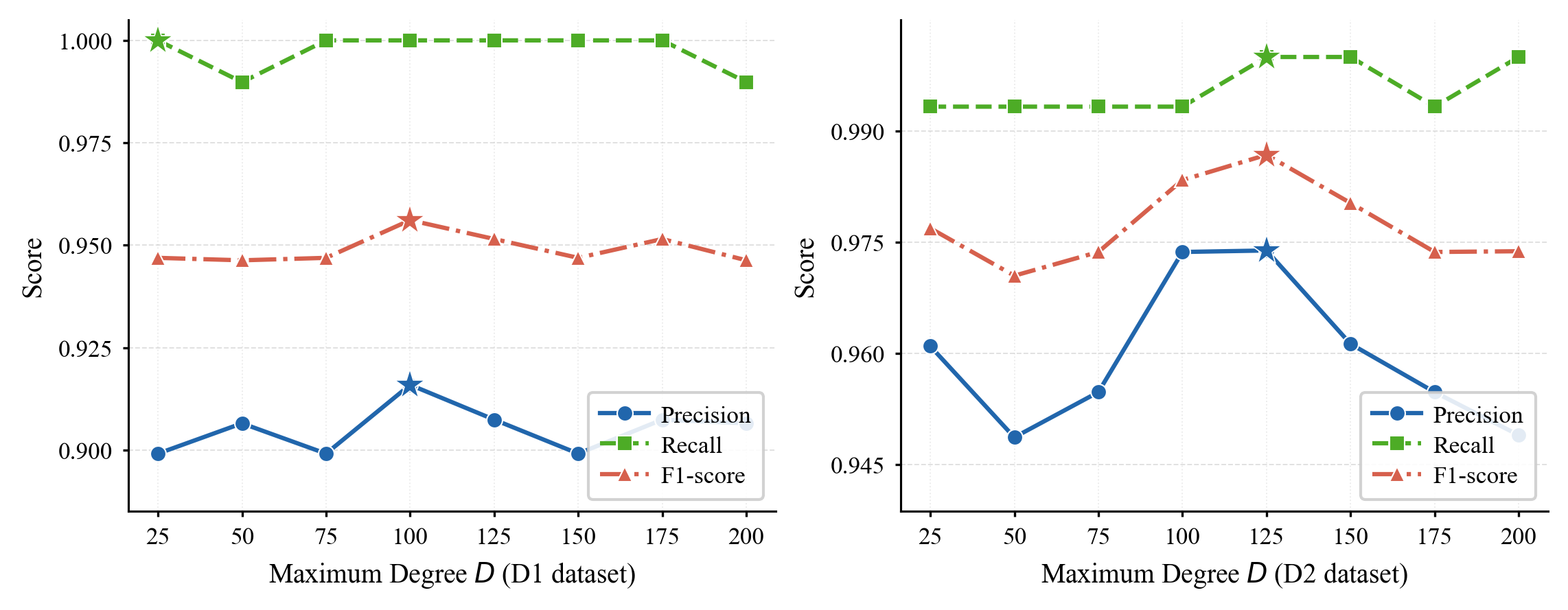}
\end{minipage}}\\
\subfigure[$N_s$ on FT task]{
\begin{minipage}[t]{0.45\textwidth}
\includegraphics[width=1\linewidth]{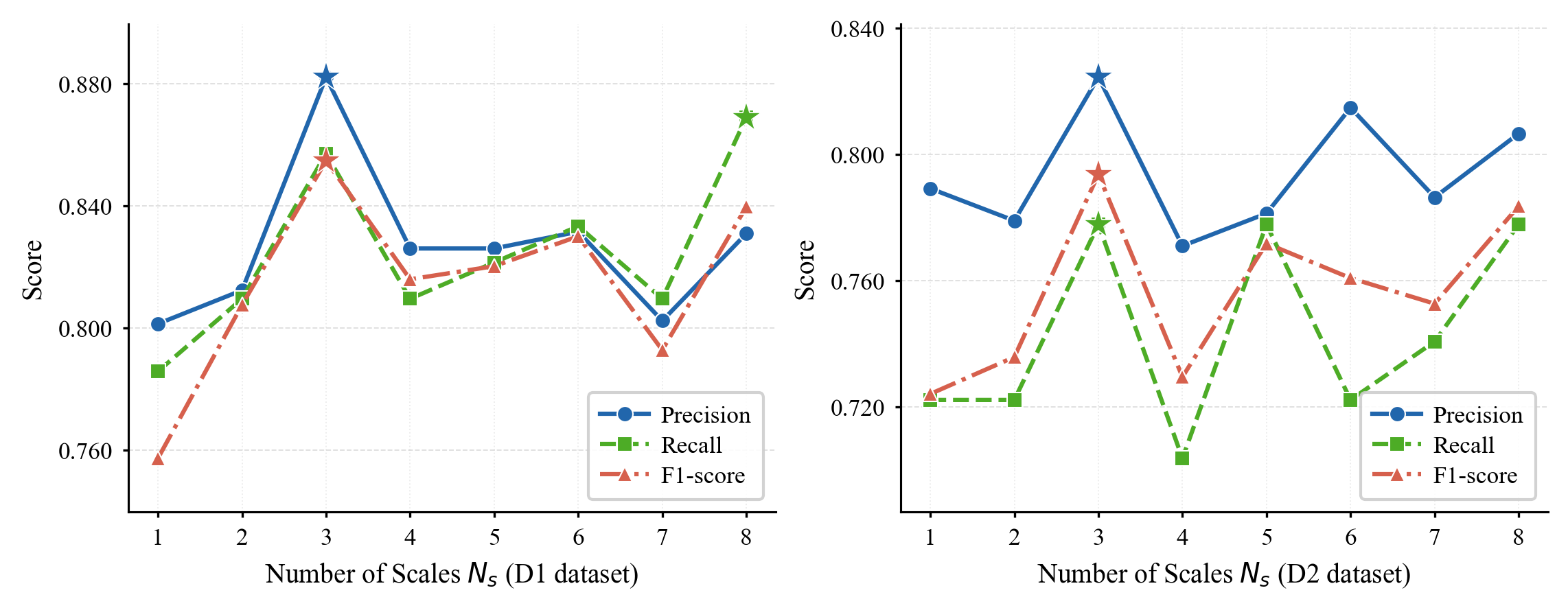}
\end{minipage}}
\subfigure[$D$ on FT task]{
\begin{minipage}[t]{0.45\textwidth}
\includegraphics[width=1\linewidth]{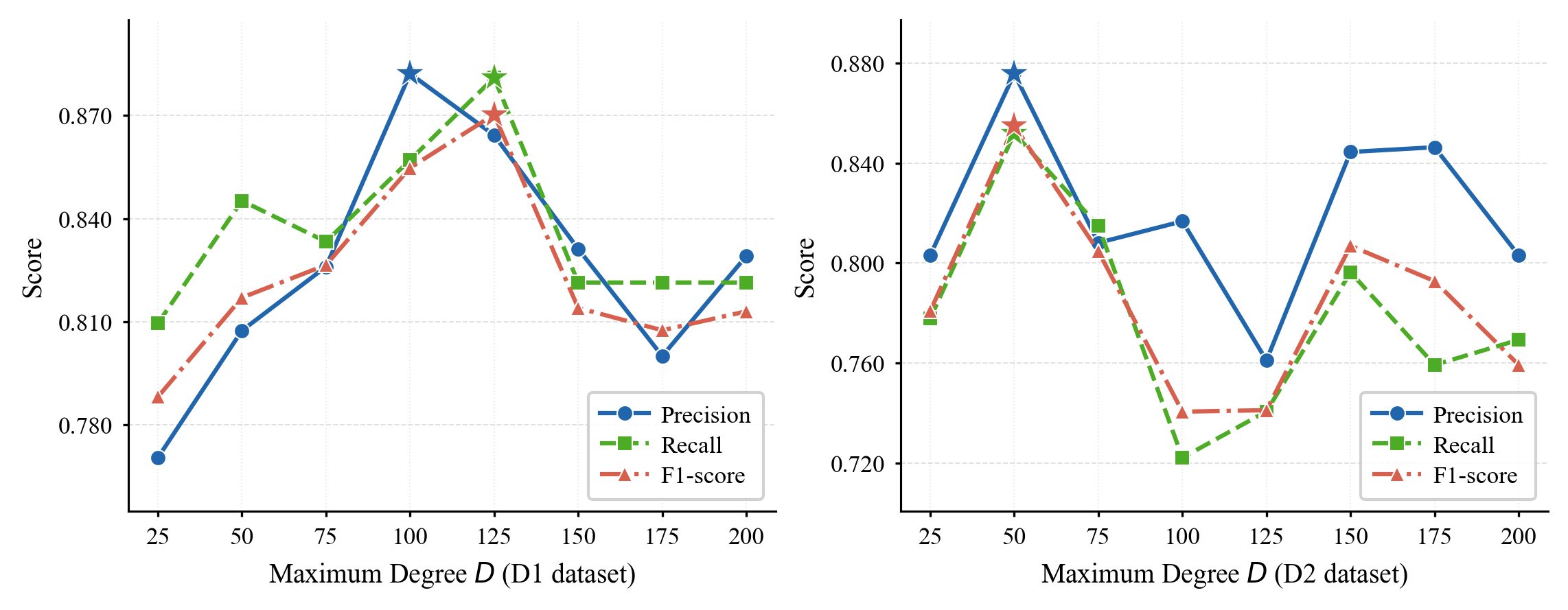}
\end{minipage}}\\
\subfigure[$N_s$ on RCL task]{
\begin{minipage}[t]{0.45\textwidth}
\includegraphics[width=1\linewidth]{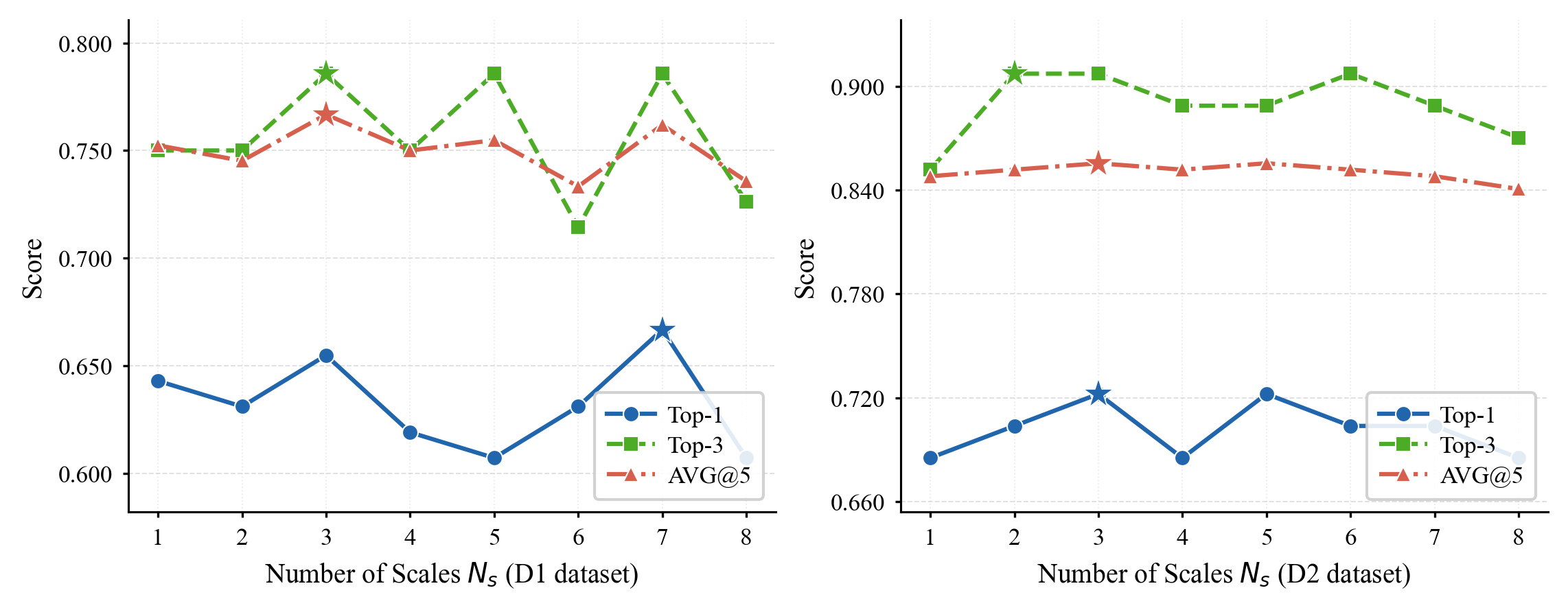}
\end{minipage}}
\subfigure[$D$ on RCL task]{
\begin{minipage}[t]{0.45\textwidth}
\includegraphics[width=1\linewidth]{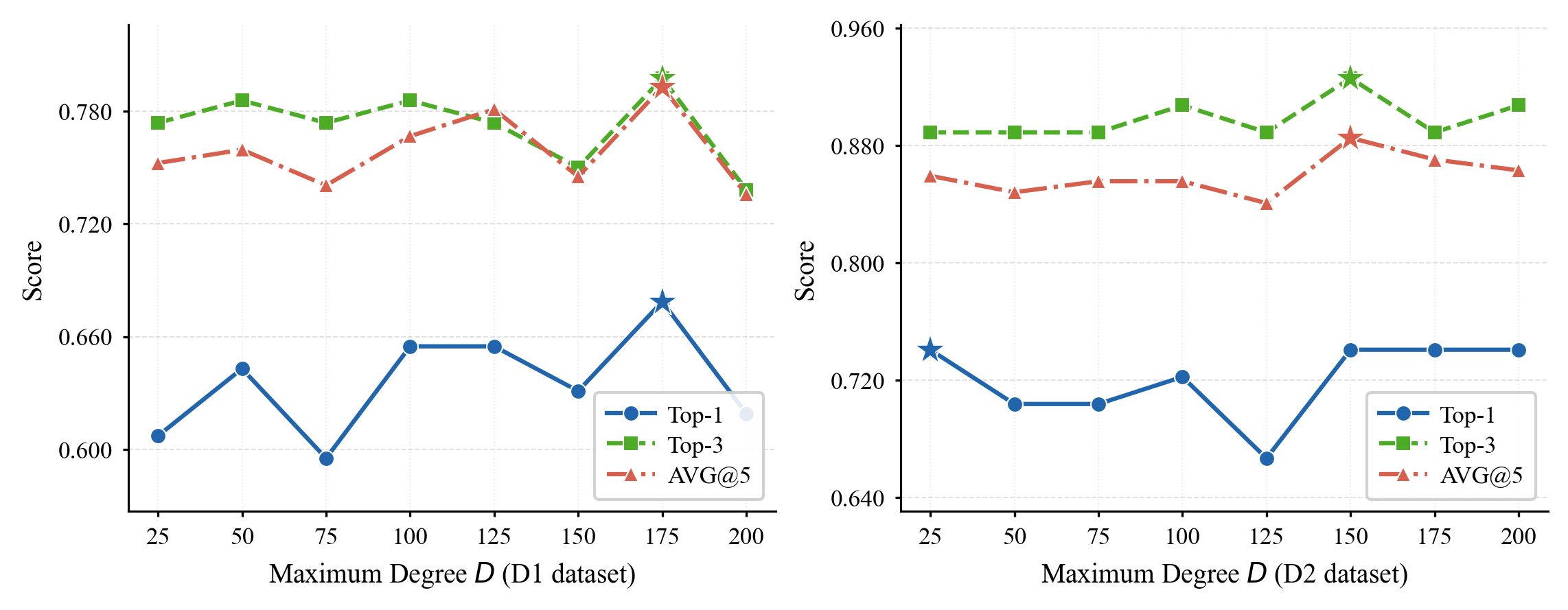}
\end{minipage}}
\caption{The experimental results for the hyperparameters number of scales ($N_s$) and maximum degree ($D$).}
\label{fig:fig7}
\end{figure*}

In this section, we conduct sensitivity analysis on two key hyperparameters in the STAR method, namely the number of scales in temporal adaptive normalization (TAN) and the maximum degree in spatial adaptive normalization (SAN). The candidate parameters are set to $N_{s}\in\{1,2,3,4,5,6,7,8\}$ with $D=100$ and $D\in\{25,50,75,100,125,150,175,200\}$ with $N_s=3$ respectively, and we show the experimental results of the hyperparameters in Figure \ref{fig:fig7}.

Regarding the hyperparameter of the number of scales $N_s$, we can find that when $N_s = 3$, the state-of-the-art (SOTA) performance is achieved in almost all cases. This is consistent with common sense because our dataset contains three modalities of data: metrics, logs, and traces. Therefore, as mentioned in Section \ref{3.2}, we keep $N_s = 3$ in our experiments.

Experimental results are highly sensitive to the maximum degree hyperparameter $D$. The optimal maximum degree varies for different tasks on different datasets. Therefore, we set specific maximum degree hyperparameters for different scenarios as shown in Table \ref{tab:D} in Appendix~\ref{imple}.

\section{Conclusion}

In this paper, we propose STAR, a spatial-temporal adaptive representation learning framework for automated incident management in microservice systems. By incorporating Temporal Adaptive Normalization (TAN) and Spatial Adaptive Normalization (SAN), STAR captures non-stationary temporal dynamics and heterogeneous service dependencies, enabling unified and explainable unsupervised learning for anomaly detection (AD), failure triage (FT), and root cause localization (RCL). Experiments on two real-world benchmarks show that STAR consistently outperforms existing baselines across all three tasks, while maintaining strong robustness, generalization, and scalability. These findings suggest that adaptive normalization offers a principled and practical foundation for robust multimodal representation learning in complex systems. Additional insights into model efficiency are provided in Appendix \ref{Model_efficiency}. Limitations and future directions are discussed in Appendix \ref{appendix:discussion}. In compliance with the single-blind review policy, the appendix, source code, and datasets for this study have been uploaded to the supplementary materials to ensure the reproducibility of this paper.

\begin{credits}
\subsubsection{\ackname}
This work was supported by grants from the National Key Research and Development Program of China (2024YFB4505903).
\end{credits}

%
%
%
\bibliographystyle{splncs04}
\bibliography{mybibliography}

\appendix

\section{Preliminaries}\label{appendix:p}
\subsection{Existing Methods}
ART (Anomaly detection, failure triage, and root cause localization) \cite{ART} is an end-to-end unsupervised event management framework for microservice systems. Its core concept is to extract shared knowledge from multimodal data to uniformly support the three key tasks of AD, FT, and RCL. The ART framework consists of three modules: dependency-aware state learning, unified fault representation acquisition, and an unsupervised solution for diagnosis tasks.
\subsubsection{Dependency-Aware Status Learning}
ART uses self-supervised learning to model three types of dependencies in a microservice system: channel dependency (CHA), time dependency (TEM), and call dependency (CAL) to predict the system's behavior under normal conditions.
\paragraph{Multimodal data serialization.}
Convert metrics, logs, and traces into a unified multi-channel time series $X^{(t)}\in\mathbb{R}^{N\times K}$, where $N$ is the number of instances and $K$ is the number of channels.
\paragraph{Multi-dependency feature extraction.}
\begin{itemize}[label=$\bullet$, leftmargin=1em]
    \item Channel dependency (CHA): Transformer encoder is used to model the relationship between channels. The self-attention mechanism is calculated as follows:
    \begin{equation}\mathrm{head}_l=\mathrm{Attention}(X^{\prime{(t)}}W_l^Q,X^{\prime{(t)}}W_l^K,X^{\prime{(t)}}W_l^V)\end{equation}
    where $X^{\prime{(t)}}\in\mathbb{R}^{K\times N}$ is the transposed input, and $W_l^Q$, $W_l^K$ and $W_l^V$ are linear projection weights.
    \item Temporal Dependence (TEM): Modeling the dynamics of time series using GRU:
    \begin{equation}h^{(t)}=\mathrm{GRU}(H^{(t)},h^{(t-1)})\end{equation}
    where $h^{(t)}$ and $h^{(t-1)}$ are the hidden states and $H^{(t)}$ is the sequence of updated snapshots.
    \item Call Dependency (CAL): Use GraphSAGE to model the call relationship between instances:
    \begin{equation}F_i^{(k)}=\mathrm{Norm}\left(\sigma\left(W^{(k)}\cdot\mathrm{Concat}(F_i^{(k-1)},F_{\mathcal{N}(i)}^{(k)})\right)\right)\end{equation}
    where $F_i^{(k)}$ is the features of instance $i$ of the $k^{th}$ layer, $\sigma$ is the Leaky ReLU activation function, $W^{(k)}$ is the learnable weight matrix of the $k^{th}$ layer, and ${\mathcal{N}(i)}$ is the set of neighbors of instance $i$.
\end{itemize}

\subsubsection{Unified Failure Representation Acquisition}
ART constructs a fault representation with clear semantic information by calculating the deviation between the predicted value and the actual observed value.
\par The deviation matrix is calculated as follows:
\begin{equation}D_{ij}=\mathrm{Agg}\left(\left(X_{ij}^{(t)}-\hat{X}_{ij}^{(t)}\right)^2\right)\end{equation}
where $t=1,2,\ldots,T$ is the window, $Agg$ is the aggregation operation and $D_{ij}$ is the aggregated deviation of instance $i$ on channel $j$.
For instance-level deviation (ILD), each row represents the deviation vector of an instance on all channels $ILD_i\in\mathbb{R}^K$.
\par For system-level deviation (SLD), it is obtained by weighted summing of the deviations of highly abnormal instances:
\begin{equation}SLD=\sum_{i\in Q}\left(\|ILD_i\|_1\times ILD_i\right)\end{equation}
where $||\cdot||_1$ is the L1-norms.
\subsubsection{Unsupervised Solutions for Diagnostic Tasks}
Then, ART uses $ILD$ and $SLD$ to complete the diagnosis tasks in an unsupervised manner.
\begin{itemize}[label=$\bullet$, leftmargin=1em]
    \item Anomaly Detection (AD): Uses Extreme Value Theory (EVT) to set a threshold for $\|SLD\|_1$ to detect system anomalies.
    \item Failure Triage (FT): Constructs a cut-tree based on the $SLD$ to perform unsupervised clustering of fault types.
    \item Root Cause Localization (RCL): Calculates the cosine similarity between the $ILD$ and the $SLD$ of each instance as a suspicious score, sorts them, and locates the root cause instance.
\end{itemize}
\par The above content briefly introduces the core ideas of ART. Please refer to the original paper \cite{ART} for details.

\subsection{Design Motivation}
\subsubsection{Motivation for Temporal Adaptive Normalization (TAN)}
\label{subsec:tan_motivation}
\paragraph{\textbf{Why multi-scale temporal analysis?}}
Microservice telemetry contains three      
modalities—metrics, logs, and traces—that operate at different temporal       
resolutions. Metric anomalies (e.g., CPU spikes) manifest at fine granularity 
(seconds), log bursts appear at medium granularity (minutes), and trace-level 
latency degradation develops over coarser windows. A single-scale             
normalization window cannot simultaneously preserve all three types of        
deviations. We therefore set $N_s=3$, one scale per modality, so that TAN's   
parameter generator receives context from all three temporal regimes before   
computing $\gamma_1, \delta_1, \gamma_2, \delta_2$.   

\paragraph{\textbf{Why two-step normalization with learnable fusion?}}
The design of TAN involves three stages: base normalization, adaptive
transformation, and weighted fusion. Each stage serves a distinct purpose:

\begin{enumerate}[leftmargin=*,noitemsep,topsep=2pt]
  \item \textbf{Base normalization} $\hat{X}_{\mathrm{base}} = (X - \mu) /
\sigma$ stabilizes training by removing global mean and variance, preventing
gradient explosion in early epochs and ensuring numerical stability across
diverse telemetry ranges (e.g., CPU percentages vs.\ millisecond latencies).

  \item \textbf{Adaptive transformation} $\hat{X}_{\mathrm{adapt}} =
\hat{X}_{\mathrm{base}} \odot \gamma_1 + \delta_1$ injects multi-scale incident
context. The parameters $\gamma_1$ and $\delta_1$ are dynamically generated from
the multi-scale features (Eq.~\ref{eq:tan_params}), allowing the normalization
to adapt to distribution shifts caused by failures. For example, during a memory
leak incident, $\gamma_1$ may amplify memory-related features while $\delta_1$
shifts the baseline to account for the new operating regime.

  \item \textbf{Weighted fusion} $\hat{X} = \alpha \odot
\hat{X}_{\mathrm{base}} + (1-\alpha) \odot \hat{X}_{\mathrm{adapt}} + \beta$
enables smooth interpolation between conservative (base) and expressive
(adaptive) normalization. This is critical because:
  \begin{itemize}[leftmargin=*,noitemsep]
      \item In \emph{stable periods} (no incident), the base normalization
suffices and the model learns $\alpha \to 1$, avoiding over-fitting to noise.
      \item In \emph{incident periods}, the adaptive branch captures the
distribution shift and $\alpha$ decreases, allowing failure-specific patterns to
dominate.
  \end{itemize}
\end{enumerate}

\subsubsection{Motivation for Spatial Adaptive Normalization (SAN)}
\label{subsec:san_motivation}
\paragraph{\textbf{Why degree-aware spatial normalization?}}
In microservice call graphs, nodes (services) exhibit heterogeneous structural
roles: hub services (high in-degree) aggregate requests from many upstream
services and are prone to bottlenecks, while leaf services (low in-degree)
handle specialized tasks and are sensitive to localized failures. Standard graph
normalization methods (e.g., BatchNorm, LayerNorm) treat all nodes uniformly,
ignoring these structural differences. SAN addresses this by embedding discrete
in-degrees into a continuous latent space via a learnable embedding $e_i =
\text{DegreeEmbedding}(d_i)$, where $d_i$ is the in-degree of node $i$. The
degree-aware correction terms $w_{\mathrm{deg}}$ and $b_{\mathrm{deg}}$
(Eq.~\ref{eq:san_degree}) are derived from the \emph{full degree distribution}
of the current graph snapshot, making them adaptive to topology changes at each
forward pass (e.g., service scaling, dynamic routing).

\paragraph{\textbf{Distinction from existing methods.}}                     
GraphNorm~\cite{GraphNorm} normalizes across all nodes in a graph with a      
single shared learnable shift, correcting for graph-level aggregation bias    
but                                                                             
 ignoring per-node structural roles. DegNorm~\cite{DegNorm} applies           
degree-based correction as a post-hoc regularizer to improve spectral         
expressiveness, without integrating it into the normalization of feature      
representations. Our SAN differs in three aspects: (1) we explicitly embed    
discrete in-degrees into a continuous latent space via a learnable embedding  
$e_i = \text{DegreeEmbedding}(d_i)$, capturing non-linear degree semantics    
beyond a simple scalar correction; (2) $w_\text{deg}$ and $b_\text{deg}$ are  
derived from the \emph{full degree distribution} of the current graph         
snapshot, making them adaptive to topology changes at each forward pass; (3)  
SAN is jointly trained with TAN under multi-task supervision (AD+FT+RCL), so  
degree embeddings are shaped by downstream incident-diagnostic objectives     
rather than unsupervised graph reconstruction.     

\subsubsection{Joint Temporal-Spatial Normalization}
\label{subsec:joint_normalization}

\paragraph{\textbf{Why combine TAN and SAN?}}
Microservice failures exhibit both temporal evolution (e.g., gradual memory
leak) and spatial propagation (e.g., cascading timeouts across dependent
services). Normalizing only the temporal dimension (e.g., using RevIN or
standard z-score) ignores the structural context of where the failure originates
and how it spreads. Conversely, normalizing only the spatial dimension (e.g.,
using GraphNorm) discards the temporal dynamics of how the failure develops over
time. Our unified framework applies TAN to temporal features (before the
Transformer encoder) and SAN to spatial features (after the GNN encoder),
ensuring that both dimensions are adaptively normalized. The residual
connections in both modules (Eq.~\ref{eq:fusion} and
Eq.~\ref{eq:san_residual}) preserve the original signal when normalization is
unnecessary, preventing over-correction in stable periods.

\section{Implementation details}
\label{appendix:i}
\subsection{Dataset Details}
\label{appendix:datasets}
We evaluate STAR on the same two real-world microservice datasets used in ART \cite{ART} to ensure fair comparison and experimental reproducibility. The datasets, denoted as \textbf{D1} and \textbf{D2}, are collected from two representative microservice systems with different architectures and operational characteristics. Both datasets contain multimodal observability data, including metrics, logs, and traces, and have been widely used as benchmarks for automated incident management.

\subsubsection{Dataset D1}
D1 is collected from a simulated e-commerce microservice system deployed in a real cloud environment with traffic patterns consistent with real business workloads. The system consists of 46 instances, including 40 microservice instances and 6 virtual machines. Failure cases were collected by replaying real-world failure scenarios over several days. The failure types include \emph{Container Hardware}, \emph{Container Network}, \emph{Node CPU}, \emph{Node Disk}, and \emph{Node Memory} failures. All failure cases are annotated with their corresponding root cause instances and failure types.

\subsubsection{Dataset D2}
D2 is collected from the production management system of a large commercial bank and consists of 18 instances, including microservices, servers, databases, and containerized components. Failure records were collected over a six-month period and manually labeled by experienced operators. The failure types include \emph{Memory}, \emph{CPU}, \emph{Network}, \emph{Disk}, \emph{JVM-Memory}, and \emph{JVM-CPU} related failures. Due to confidentiality constraints, D2 is not publicly available and is used under a non-disclosure agreement. D2 has also been used in the International AIOps Challenge.

\subsubsection{Data Characteristics and Distribution Shift}
Table~\ref{tab:table1} summarizes the key statistics of the two datasets, including the number of instances, failure cases, normal cases, failure types, and the scale of each data modality. The substantial volume and diversity of multimodal data make these datasets suitable for evaluating unified and unsupervised incident management methods.

\begin{figure}[h]
\centering
\includegraphics[width=\linewidth]{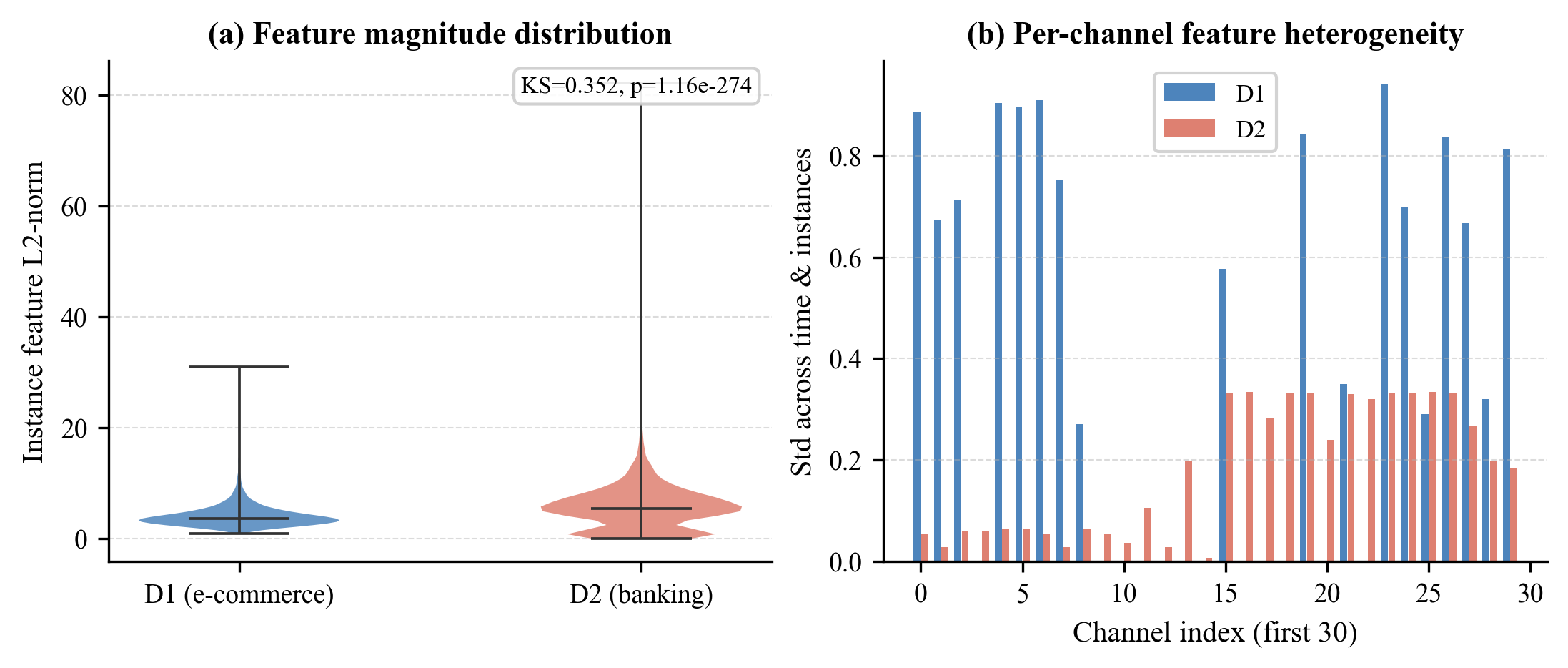}
\caption{Feature magnitude distribution (violin) and per-channel feature
heterogeneity (bar) for D1 (e-commerce) and D2 (banking). KS test confirms a
statistically significant distributional shift between datasets ($p <
0.001$).}
\label{fig:dist_feat}
\end{figure}

\begin{figure}[h]
\centering
\includegraphics[width=\linewidth]{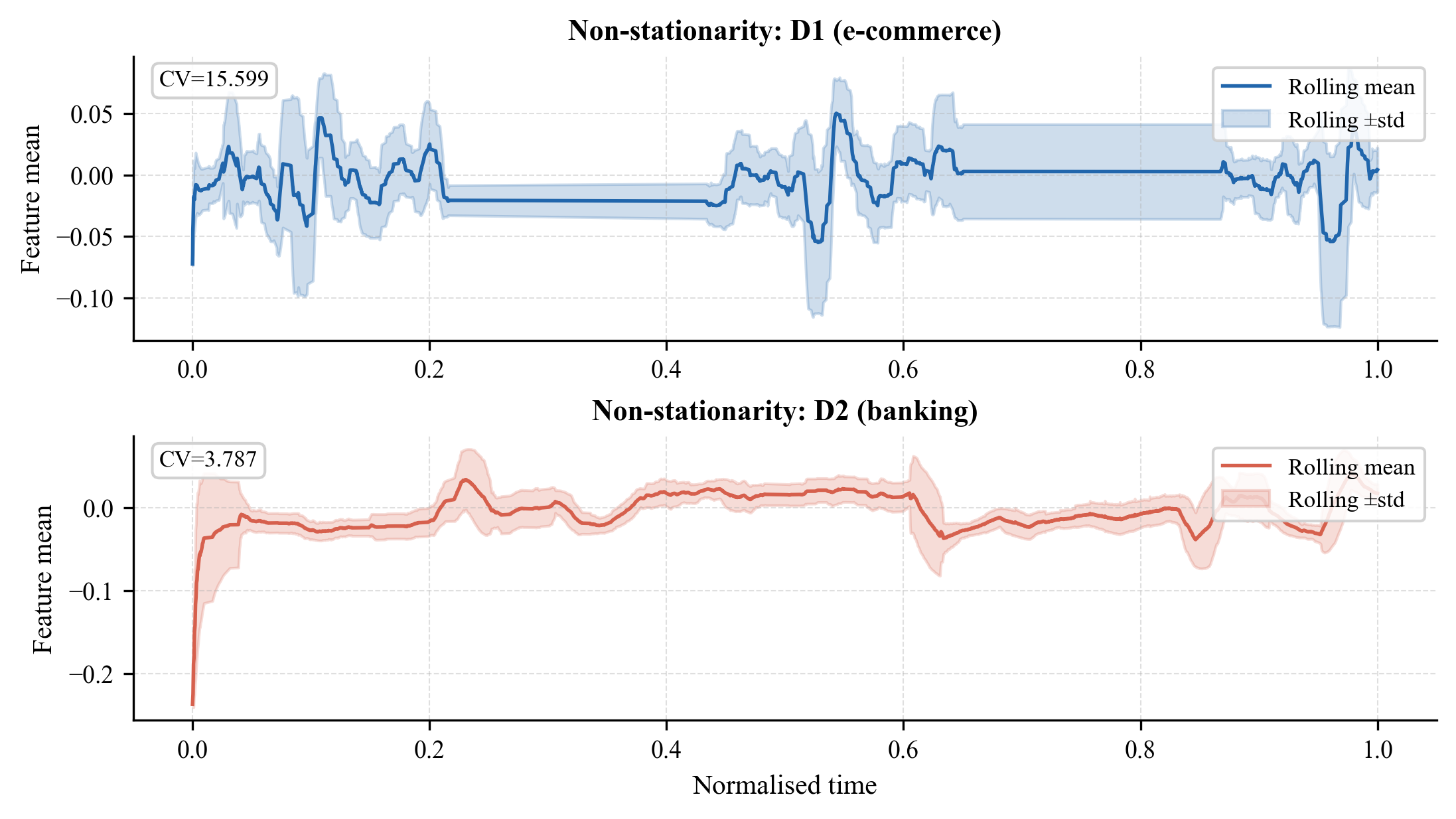}
\caption{Rolling mean and standard deviation of instance feature norms over
normalized time. D2 exhibits markedly higher temporal variation (CV),
confirming the motivation for Temporal Adaptive Normalization (TAN).}
\label{fig:dist_nonstat}
\end{figure}

\begin{figure}[h]
\centering
\includegraphics[width=\linewidth]{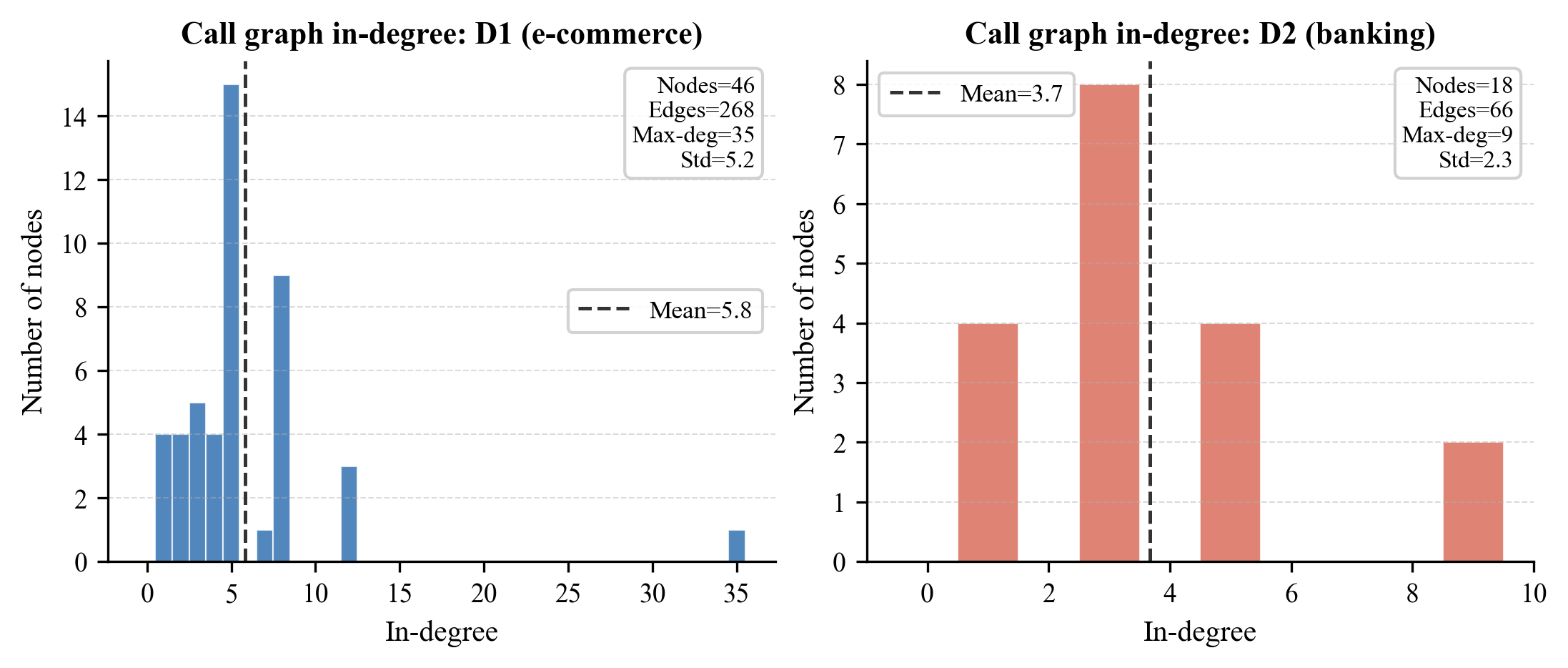}
\caption{In-degree distributions of the service call graphs in D1
(e-commerce, left) and D2 (banking, right). Both datasets exhibit
heterogeneous topologies with skewed degree distributions. D2 shows a
heavier tail and higher degree variance, motivating degree-aware Spatial
Adaptive Normalization (SAN) to prevent high-degree hub nodes from
dominating graph-level statistics. Dashed lines indicate the mean in-degree;
inset statistics report node/edge counts, maximum degree, and degree standard
deviation.}
\label{fig:dist_indegree}
\end{figure}

Figure~\ref{fig:dist_feat} ,~\ref{fig:dist_nonstat}, and ~\ref{fig:dist_indegree} characterize the two datasets along three
complementary axes.
\textit{(a) Feature magnitude:} D1 and D2 exhibit substantially different
feature-norm distributions (KS test $p < 0.001$), confirming the need for
dataset-adaptive normalization (Figure~\ref{fig:dist_feat}).
\textit{(b) Non-stationarity:} The rolling mean and standard deviation of D2
vary markedly over time (higher coefficient of variation, CV), motivating the
use of TAN to capture time-varying statistics (Figure~\ref{fig:dist_nonstat}).
\textit{(c) Graph topology:} D1 has a more uniform in-degree distribution
while D2 shows a heavier tail, justifying degree-aware SAN to prevent
high-degree hubs from dominating graph-level statistics
(Figure~\ref{fig:dist_indegree}).

\subsubsection{Train-Test Split}
Following the protocol in ART \cite{ART}, we split the datasets based on failure timestamps. Specifically, the earliest timestamp among the last 40\% of failure cases is used as the split point. Data before this point are used for model training and initialization, while the remaining data are used for evaluation. This temporal split avoids information leakage and reflects realistic deployment scenarios.

\subsection{Metrics Details}
\label{appendix:metrics}
We evaluate the performance of STAR on three diagnostic tasks: anomaly detection (AD), failure triage (FT), and root cause localization (RCL), following standard evaluation protocols in prior work.

\paragraph{Anomaly Detection and Failure Triage.}
Anomaly detection (AD) is formulated as a binary classification task that determines whether a system failure occurs, while failure triage (FT) is formulated as a multi-class classification task that predicts the failure type. For both tasks, we report \emph{Precision}, \emph{Recall}, and \emph{F1-score}, which are defined as:
\begin{equation}
\text{Precision} = \frac{TP}{TP + FP}, \quad
\text{Recall} = \frac{TP}{TP + FN},
\end{equation}
\begin{equation}
\text{F1-score} = \frac{2 \times \text{Precision} \times \text{Recall}}{\text{Precision} + \text{Recall}},
\end{equation}
where $TP$, $FP$, and $FN$ denote the numbers of true positives, false positives, and false negatives, respectively.

Since the failure types in FT are imbalanced, we additionally report the \emph{weighted F1-score}, where the F1-score of each class is weighted by its support in the evaluation set, following standard practice~\cite{Weighted}.

\paragraph{Root Cause Localization.}
Root cause localization (RCL) aims to identify the faulty instance responsible for a failure. We adopt \emph{Top-$k$ accuracy} as the evaluation metric, which measures whether the ground-truth root cause appears among the top-$k$ predicted candidates:
\begin{equation}
\text{Top-}k = \frac{1}{N} \sum_{i=1}^{N} \mathbb{I}\bigl(g_i \in \mathrm{Top}_k(P_i)\bigr),
\end{equation}
where $N$ is the number of evaluated failure cases, $g_i$ denotes the ground-truth root cause of the $i$-th failure, $P_i$ is the ranked list of predicted root causes, and $\mathbb{I}(\cdot)$ is the indicator function.

To provide a more stable and comprehensive assessment, we further report the average Top-$k$ accuracy over $k \in \{1,2,3,4,5\}$:
\begin{equation}
\text{AVG@5} = \frac{1}{5} \sum_{k=1}^{5} \text{Top-}k.
\end{equation}

\subsection{Implementation details}\label{imple}
We implement STAR in PyTorch and DGL framework, using the same experimental settings, training strategies, and hyperparameter ranges as the ART method \cite{ART}. Specifically for the STAR method, the number of scales in temporal adaptive normalization (TAN) is set to $N_s = 3$, and the parameter setting of the maximum degree in spatial adaptive normalization (SAN) is shown in Table \ref{tab:D}.

\begin{table*}[h]
\caption{The hyperparameter setting table of maximum degree $D$.}
\centering
\label{tab:D}
\begin{tabular}{l|lll|lll}
\hline
Datasets         & \multicolumn{3}{l|}{D1}                                   & \multicolumn{3}{l}{D2}                                   \\ \hline
Task             & \multicolumn{1}{l|}{AD}  & \multicolumn{1}{l|}{FT}  & RCL & \multicolumn{1}{l|}{AD}  & \multicolumn{1}{l|}{FT} & RCL \\ \hline
Maximum Degree $D$ & \multicolumn{1}{l|}{100} & \multicolumn{1}{l|}{125} & 175 & \multicolumn{1}{l|}{125} & \multicolumn{1}{l|}{50} & 150 \\ \hline
\end{tabular}
\end{table*}

\section{Model Efficiency}\label{Model_efficiency}
STAR is designed to satisfy the efficiency requirements of large-scale, latency-sensitive microservice environments. Although temporal adaptive normalization (TAN) and spatial adaptive normalization (SAN) introduce additional computation, both modules are lightweight and highly parallelizable. TAN relies on shallow multi-scale temporal convolutions, and SAN conditions normalization on simple structural features, avoiding expensive graph-level operations.

Table~\ref{tab:table7} reports the end-to-end runtime of STAR for anomaly detection (AD), failure triage (FT), and root cause localization (RCL) on the D1 and D2 datasets. STAR achieves efficient inference across all tasks, with runtimes well within real-time or near-real-time diagnostic requirements.

\begin{table*}[h]
\caption{End-to-end runtime (in seconds) of STAR for each task on the D1 and D2 datasets.}
\centering
\label{tab:table7}
\begin{tabular}{l|lll|lll}
\hline
Datasets    & \multicolumn{3}{l|}{D1}                                        & \multicolumn{3}{l}{D2}                                         \\ \hline
Tasks       & \multicolumn{1}{l|}{AD}    & \multicolumn{1}{l|}{FT}   & RCL   & \multicolumn{1}{l|}{AD}    & \multicolumn{1}{l|}{FT}   & RCL   \\ \hline
Runtime (s) & \multicolumn{1}{l|}{12.52} & \multicolumn{1}{l|}{7.64} & 23.73 & \multicolumn{1}{l|}{27.28} & \multicolumn{1}{l|}{3.50} & 10.09 \\ \hline
\end{tabular}
\end{table*}

Overall, STAR balances representation quality and computational cost, enabling practical end-to-end deployment.

\section{Visualization of Adaptive Normalization Effects}
\label{appendix:tsne}
To provide intuitive evidence that TAN and SAN transform the feature space
into a more structured and discriminative representation, we apply
t-SNE to intermediate features extracted at four hook points: raw
input (before TAN), post-TAN normalized features, pre-SAN GNN output, and
post-SAN output. Features are colored by fault type; gray points represent
normal windows.

\begin{figure}[h]
\centering
\includegraphics[width=\linewidth]{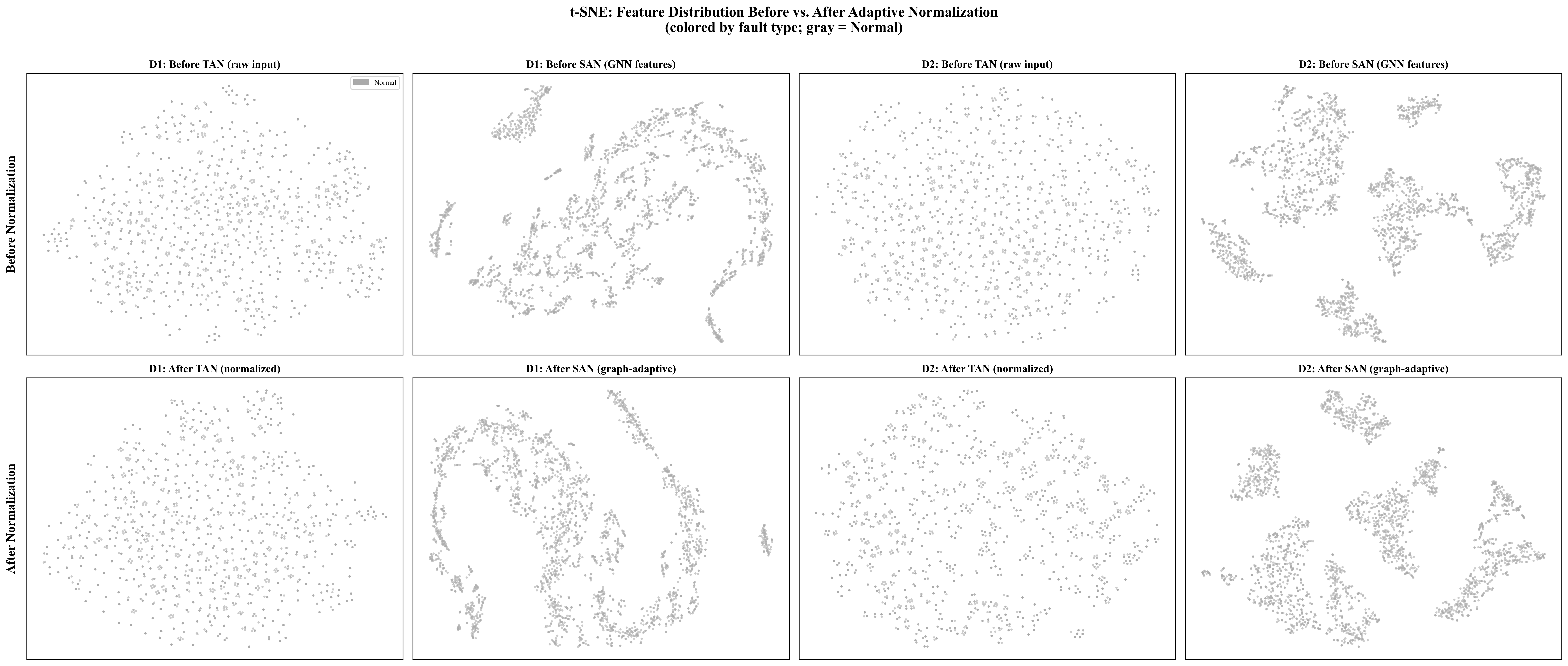}
\caption{t-SNE visualization of intermediate representations before and after
Temporal Adaptive Normalization (TAN, left two columns) and Spatial Adaptive
Normalization (SAN, right two columns) on D1 and D2. After normalization,
fault-type clusters (colored) become more compact and better separated from
normal traffic (gray), validating that TAN and SAN produce more discriminative
representations.}
\label{fig:tsne}
\end{figure}

As shown in Figure~\ref{fig:tsne}, after TAN the temporal feature clusters
become more compact, reducing the overlap between fault types and normal
states. After SAN, instances sharing similar graph structural roles cluster
together, improving the separability of root cause instances. These
observations directly corroborate the ablation results in
Table~\ref{tab:table6} and confirm that adaptive normalization produces
qualitatively richer representations beyond what static normalization
provides.

\section{Limitations and Future Directions}
\label{appendix:discussion}
While STAR achieves strong and consistent performance across anomaly detection (AD), failure triage (FT), and root cause localization (RCL), several limitations suggest directions for future research.

\subsection{Scalability and Computational Efficiency}
STAR introduces Temporal Adaptive Normalization (TAN) and Spatial Adaptive Normalization (SAN) to enhance robustness under non-stationary temporal dynamics and heterogeneous service dependencies. Although both components are lightweight and highly parallelizable, they introduce additional computation compared to static normalization. As shown in Table~7 in the main paper, STAR meets real-time or near-real-time inference requirements on the evaluated benchmarks. Nevertheless, further optimization may be necessary for ultra-large-scale systems or resource-constrained edge environments. Future work may explore efficiency-oriented techniques such as model pruning~\cite{MC1}, knowledge distillation~\cite{KD1}, and dynamic inference~\cite{DI1,DI2} to reduce computational and memory overhead.

\subsection{Dynamic Topology Modeling}
STAR captures service interactions through graph-based representations with structure-aware normalization. While SAN alleviates sensitivity to degree imbalance, the current framework assumes relatively stable service topologies within short time windows. In environments with frequent instance creation, removal, or migration, rapid topology changes may affect dependency modeling and root cause localization. Incorporating dynamic graph learning methods, such as Dynamic Graph Neural Networks~\cite{DGNN1,DGNN2} or Temporal Graph Networks~\cite{TGN1}, is a promising direction to explicitly model evolving service dependencies.

\subsection{Adaptation to Evolving Failure Patterns}
The failure triage module in STAR leverages historical failure patterns to initialize the clustering structure, which may limit adaptability to previously unseen failure types during early deployment. Integrating online or incremental learning mechanisms~\cite{OL1} could enable continuous adaptation to evolving system behaviors and emerging failure patterns without full retraining.

Overall, these limitations do not detract from the core contributions of STAR, but rather point to opportunities for extending adaptive normalization and self-supervised representation learning toward more scalable, adaptive, and resilient incident management systems.

\end{document}